\documentclass[runningheads,a4]{llncs}
\newif\ifdraft
\newcommand{\silcorcomment}[1]{\ifdraft{\leavevmode\color{red}{[SC]: 
  {#1}}}\else{\vspace{0ex}}\fi}

\newcommand{\fabsebcomment}[1]{\ifdraft{\leavevmode\color{cyan}{[FS]: 
  {#1}}}\else{\vspace{0ex}}\fi}
\newcommand{\sout}[1]{\ifdraft{\sout{#1}}\else{\vspace{0ex}}\fi}
\usepackage[sort&compress]{natbib}  
\usepackage[lined,boxed,linesnumbered]{algorithm2e}
\usepackage{amsmath}
\usepackage{cite}
\usepackage[english]{babel}
\usepackage{multibib}
\usepackage{multicol}  
\usepackage{multirow}  
\usepackage{pifont}
\usepackage{rotating}  
\usepackage{soul}  
\usepackage[table]{xcolor}
\usepackage[utf8]{inputenc} 
\usepackage[normalem]{ulem} 
\usepackage{url} 
\newcites{prim}{Primary sources}
\newcites{sec}{References}
\newcommand{\TP}{\mathrm{TP}}

\newcommand{\FP}{\mathrm{FP}}
\newcommand{\FN}{\mathrm{FN}}

\newcommand{\cmark}{\ding{51}}%
\newcommand{\xmark}{\ding{55}}

\newcommand{\killpunct}[1]{} 
\newcolumntype{.}{D{.}{.}{-1}}
\newcolumntype{M}[1]{>{\centering\arraybackslash}m{#1}}
\newcolumntype{N}{@{}m{0pt}@{}}
\newcolumntype{C}[1]{>{\centering\let\newline\\\arraybackslash\hspace{0pt}}m{#1}}
\newcolumntype{Y}{>{\centering\arraybackslash}X}
\begin{document}
%
% \title{Syllabic Quantity Patterns as Rhythmic Features for Latin
% Authorship Attribution}
\title{Syllabic Quantity Patterns \\ as Rhythmic Features \\ for Latin
Authorship Attribution}

\titlerunning{Syllabic Quantity for Latin Authorship Attribution}

\author{Silvia Corbara$^{1}$, %\orcidID{0000-0002-5284-1771}
Alejandro Moreo$^{2}$, %\orcidID{0000-0002-0377-1025}
Fabrizio Sebastiani$^{2}$ %\orcidID{0000-0003-4221-6427}
}

\authorrunning{Silvia Corbara, Alejandro Moreo, Fabrizio Sebastiani}

\institute{$^{1}$Scuola Normale Superiore \\ 56126 Pisa (IT) \\
\email{silvia.corbara@sns.it} \\
$^{2}$Istituto di Scienza e Tecnologie dell'Informazione \\
Consiglio Nazionale delle Ricerche \\
56124 Pisa, Italy \\
\email{firstname.lastname@isti.cnr.it} }

\maketitle

\begin{abstract}
  It is well known that, within the Latin production of written text,
  peculiar metric schemes were followed not only in poetic
  compositions, but also in many prose works. Such metric patterns
  were based on so-called \textit{syllabic quantity}, i.e., on the
  length of the involved syllables, and there is substantial evidence
  suggesting that certain authors had a preference for certain metric
  patterns over others. In this research we investigate the
  possibility to employ syllabic quantity as a base for deriving
  rhythmic features for the task of computational authorship
  attribution of Latin prose texts. We test the impact of these
  features on the authorship attribution task when combined with other
  topic-agnostic features. Our experiments, carried out on three
  different datasets, using two different machine learning methods,
  show that rhythmic features based on syllabic quantity are
  beneficial in discriminating among Latin prose authors.
\end{abstract}

% --------------------------------------------------------------------

% \tableofcontents

\section{Introduction}
\label{sec:intro}

\noindent In the study of textual documents, \textit{Authorship
Analysis} can be defined ``broadly as any attempt to infer the
characteristics of the creator of a piece of linguistic data''
\citep[p.\ 238]{Juola:2006jn}, where these characteristics include the
author’s biographical information (e.g., age, gender, mother tongue,
etc.) and identity. In particular, the set of tasks grouped under the
name of \textit{Authorship Identification} (AId) concern the study of
the true identity of the author of a text when it is unknown or
debated. The three main tasks of AId are \textit{Authorship
Attribution} (AA), \textit{Authorship Verification} (AV), and
\textit{Same-Authorship Verification} (SAV).
% In AA, given a set of candidate authors $\{A_{1}, \ldots, A_{m}\}$
% and a document $d$, the goal is to find the most probable author of
% document $d$ among the set of candidates;
In AA \citep{Koppel:2009ix, Stamatatos:2009ye}, given a document $d$
and a set of candidate authors $\{A_{1}, \ldots, A_{m}\}$, the goal is
to predict the real author of $d$ among the set of candidates; AA is
thus a single-label multi-class classification problem, where the
classes are the authors in the set
$\{A_{1}, \ldots, A_{m}\}$.\footnote{In classification,
\textit{multi-class} (as opposed to \textit{binary}) means that there
is a set of $m>2$ classes to choose from; there are instead just 2
classes to choose from in the binary case. On the other hand,
\textit{multi-label} (as opposed to \textit{single-label}) means that
zero, one, or more than one class may be attributed to each item;
exactly one class must instead be attributed to any given item in the
single-label case.} In AV \citep{Stamatatos:2016ij}, given a single
candidate author $A$ and a document $d$, the goal is to infer whether
$A$ is the real author of $d$ or not; AV is thus a binary
classification problem, with $A$ and $\overline{A}$ as the possible
classes. In SAV \citep{Koppel:2014bq}, given two documents $d_{1}$ and
$d_{2}$, the goal is to infer whether the two documents $d_{1}$ and
$d_{2}$ are by the same author or not; SAV is thus also a binary
classification problem, with \textsc{Same} and \textsc{Different} as
the possible classes.

Generally speaking, the goal of AId is to find a way to spot the
``hand'' of a given writer, so as to clearly separate his/her written
production from those of other authors. Hence, the core of this
practice, also known as ``stylometry'', does not rely on the
investigation of the artistic value or the meaning of a written work,
but on a \textit{quantifiable} characterisation of its style. This
characterisation is typically achieved via an analysis of the
frequencies of linguistic events (also known as ``style markers''),
where the frequencies of these events are assumed to remain more or
less constant throughout the production of a given author (and,
conversely, to vary substantially across different authors) \citep[p.\
241]{Juola:2006jn}. This approach often relies on textual traits of
apparently minimal significance (such as the use of punctuation or
conjunctions), which however are assumed to be out of the conscious
control of the writer, and hence hard to modify or imitate. In his
essay \textit{Clues}, the noted historian Carlo
\citet{Ginzburg:1989ls} describes the emergence of this analytical
approach (which he traces back to the late 18th century) in a number
of fields of human activity, and calls it the \textit{evidential
paradigm}.

As hinted above, (computational) AId tasks are often solved by means
of a \textit{text classification} approach, in which the texts of
unknown authorship are the objects of classification and the classes
represent authors (as in AA or AV) or same/different authorship (as in
SAV). In turn, text classification is usually solved via
\textit{supervised machine learning}, whereby a general-purpose
supervised learning algorithm trains a classifier to perform
authorship identification by exposing it to a set of training examples
(i.e., texts by the authors of interest and whose authorship is
certain).

In this work we focus on the AA task for Latin prose documents, and
experiment with the idea of using \textit{syllabic quantity}
\citep{Sturtevant:1922ll} in order to derive an additional set of
stylistic features for this task. Syllables are categorized as
``long'' or ``short'' based on their ``quantity'' (see
Section~\ref{sec:method_Latin_prosody}), and peculiar sequences of
long and short syllables were used by Latin authors as metric (i.e.,
rhythmic) patterns. Our idea to use these sequences as features for
performing AId is based on accumulated evidence (again, see
Section~\ref{sec:method_Latin_prosody}) which suggests that some Latin
authors show a remarkable preference, more or less conscious, for
specific rhythmic patterns obtained by specific sequences of long and
short syllables, even in prose texts.  In order to assess the
plausibility of this idea we run a number of experiments, using three
different datasets and two different learning methods, in which we
evaluate the effect of features based on the notion of ``syllabic
quantity'' on the classification when coupled with (other)
topic-agnostic features.

The rest of this paper is organized as follows. In
Section~\ref{sec:method} we first present our methodological setting,
introducing some theoretical background regarding Latin prosody and
the concept of syllabic quantity, and discussing how we extract the
latter from text. In Section~\ref{sec:exps} we present our
experimental setting, including the datasets we employ and the
experimental protocol we follow; we then discuss our results in
Section~\ref{sec:results}. In Section~\ref{sec:relatedwork} we present
some related work, while in Section~\ref{sec:conclusion} we conclude
with some final remarks and discussion of avenues for future research.

The code to reproduce all our experiments is available at:
\url{https://github.com/silvia-cor/SyllabicQuantity_Latin}.

% ----------------------------------------------------------------------

\section{Methodological setting}
\label{sec:method}

\noindent In this section we give an introduction to Latin prosody
(Section~\ref{sec:method_Latin_prosody}) as a necessary premise to our
research; in Section~\ref{sec:method_SQ_feats} we then describe the
tool we use in order to extract syllabic quantity from Latin prose
texts.

% ----------------------------------------------------------------------

\subsection{A brief introduction to Latin prosody}
\label{sec:method_Latin_prosody}

\noindent As other languages, Latin is based on \textit{syllables},
i.e., sound units a single word can be divided into, which can be
thought of as oscillations of sound in the pronunciation of the
word. Every Latin word has as many syllables as it has vowels or
diphthongs.\footnote{Diphthongs are combinations of two vowels that
count as a single, long vowel. In Latin, only the combinations ``ae'',
``au'', ``ei'', ``eu'', ``oe'', ``ui'' are diphthongs.} Generally
speaking, a Latin word is divided into syllables according to the
following rules:
\begin{itemize}

\item a single consonant and the vowel that follows it belong to the
  same syllable, e.g., ``pater'' (``father'') divides into two
  syllables as ``pa-ter'';
 
\item two adjacent consonants belong to two adjacent syllables, e.g.,
  ``mitto'' (``I send'') divides as ``mit-to'', and ``arma''
  (``weapons'') divides as ``ar-ma'';
 
\item compounds generate different syllables, e.g., ``abest''
  (``he/she/it is missing/away''), being composed of the preposition
  ``ab'' (``from'') and the verb ``est'' (``he/she/it is''), divides
  as ``ab-est''.
 
\end{itemize}
\noindent A syllable is characterized by its \textit{quantity}, which
refers to the amount of time required to pronounce it. Specifically, a
syllable can be \textit{long} or \textit{short}, and this is
determined first and foremost by the quantity of its vowel, and then
by the consonant sounds that follow it. In fact, a single vowel has
its own quantity, which in turn depends on the structure of the word,
or on its etymology: so, for example, a vowel before another vowel
(when the two do not form a diphthong) is short, while a vowel
originating from a contraction, such as ``nil'', contracted from
``nihil'' (``nothing''), is long. In the study of syllabic quantities,
long vowels are traditionally marked with a \textit{macron} (\={a}),
while short vowels are (only sometimes) marked with a \textit{breve}
(\u{a}). A syllable is said to be \textit{short} if it contains a
short vowel; it is said to be \textit{long ``by nature''} if it
contains a long vowel (or a diphthong), and it is said to be
\textit{long ``by position''} if it contains a short vowel followed
either by two consonants or by a double consonant (``x'' or ``z'').

Note that the explanations offered here regarding both the
syllabification and the quantification rules for Latin are rather
generic, and many more detailed rules exist, with exceptions and
specific cases, making the study of prosody all but a trivial matter;
see the expositions by \citet{Ayer:2014lm, Ceccarelli:2018so,
Harrison:web, Sturtevant:1922ll}, for a more complete discussion of
these topics.

It is well known that classical Latin (and Greek) poetry followed
metric patterns based on syllabic quantity, i.e., on well-chosen
sequences of short and long syllables. In particular, syllables were
combined in what is called a \textit{foot}, and a series of feet
composed the \emph{metre} of a verse. For example, one of the most
renowned meters is the \textit{dactylic hexameter} (employed, among
others, by Virgil in his \textit{Aeneid}), which is composed of $6$
\textit{dactyl feet} (each consisting of a long and two short
syllables), with the possibility of a substitution with a
\textit{spondee} (which consists of two long syllables) in most
positions; additionally, the sixth foot can be a \textit{trochee}
(consisting of a long and a short syllable). An example of a dactylic
hexameter is\footnote{Regarding the following notation, ``$-$'' stands
for a long syllable, ``$\cup$'' stands for a short syllable, and
``$X$'' stands for an \textit{anceps}, which can be either a short or
a long syllable; the vertical bar indicates where one foot ends and
the other begins, and the double vertical bar indicates where the
dactylic hexameter ends. Concerning this example, we observe that the
particle ``-que'' is always a short syllable, and that an ``i''
between vowels has a consonant function.}
\begin{samepage}
  \begin{center}
    $-\cup\cup | -\cup\cup | - - | - - | -\cup\cup | -X ||$ \\
    Arma vi$|$rumque ca$|$n\={o}, Tr\={o}$|$iae qu\={\i}$|$
    pr\={\i}mus ab$|$ \={o}r\={\i}s$||$
  \end{center}
\end{samepage}
\noindent Similar metric schemes were followed also in many prose
compositions, in order to give a certain cadence to the discourse, and
to focus the attention on specific parts of it. In particular, the end
of sentences and periods was deemed to be especially important in this
sense, and it is known as \textit{clausula}.  Orators such as Cicero
were particularly aware of the effects of such rhythmic endings, as in
the example below (consisting of a \textit{molossus}, i.e., a foot
consisting of $3$ long syllables, followed by a \textit{cretic}, i.e.,
a foot consisting of the sequence ``long-short-long''):

\begin{samepage}
  \begin{center}
    $-\;-\;- | -\cup- ||$ \\
    c\={o}nsulum scelus, cupidit\={a}s, e\textbf{gest\={a}s,
    au$|$d\={a}cia!}$||$
  \end{center}
\end{samepage}
\noindent During the Middle Ages, Latin prosody underwent a gradual
but profound change, that also propagated to romance languages: the
concept of syllabic quantity lost relevance in favour of the
\textit{accent}, or \textit{stress}. As a matter of clarification,
both phenomena, syllabic quantity and accent, were present in both
classic and medieval Latin. However, in classic Latin stress did not
have a role in rhythmic composition (as we have seen), and was
pronounced with a higher \textit{pitch}. Instead, medieval Latin
speakers gradually stopped ``hearing'' the quantities of word
syllables, in favour of a higher \textit{intensity} given by stress,
even though a stressed syllable typically requires also a longer time
to be pronounced. The modern consequences of this process can be seen,
for example, in Italian poetry, where a verse is characterized by the
number of syllables (but not their quantities) and the positions of
the accents. Moreover, Latin accentuation rules are largely dependent
on syllabic quantity; so, for example, words longer than two syllables
are accented on the next-to-last syllable if this syllable is long,
e.g., ``am\'{\i}cus'' (``friend''), otherwise they are accented on the
third-to-last syllable, e.g., ``d\'{o}m\u{\i}nus'' (``master'').

In the middle of these transformations, medieval writers retained the
classical importance of the \textit{clausula}, although it now
followed the change in paradigm and became based on stress rather than
quantity. Stress-based rhythmic patterns are known as \textit{cursus}.
We can distinguish the following three main types of
cursus:\footnote{Regarding the following notation, ``$-$'' stands for
a stressed syllable and ``$+$'' stands for an unstressed syllable. See
for example \citep[]{Oberhelman:1984nw, Janson:1975po} for an in-depth
analysis of this stylistic technique.}

\medskip

\begin{tabular}{lll}
  \textit{Cursus planus}: 
  & \hspace{0.5em} \( - + | + - + || \) 
  & \hspace{0.5em} \'{\i}llum ded\'{u}xit \\
  \textit{Cursus tardus}: 
  & \hspace{0.5em} \( - + | + - + + || \) 
  & \hspace{0.5em} \'{\i}re tent\'{a}verit \\ 
  \textit{Cursus velox}: 
  & \hspace{0.5em} \( - + + | + + - + || \) 
  & \hspace{0.5em} sa\'{e}cula saecul\'{o}rum
\end{tabular}
 
\medskip

\noindent Many scholars (see e.g., \citep{Janson:1975po,
Keeline:2019au}) have shown that certain authors preferred specific
types of rhythmic patterns, and that differences can be detected even
between authors who cannot be assumed to consciously care for such
rhythmic patterns at all, both in metric based on quantities
\citep[p.\ 187]{Keeline:2019au} and in metric based on accents
\citep[p.\ 20]{Janson:1975po}. This is an important point:
\textit{many authors consistently show a certain (more or less
conscious) preference for specific rhythmic constructions, even if
they do not follow the prosodic canons of the time.}

An author's use of certain rhythmic patterns might thus play an
important role in the identification of that author's style; in fact,
it has already been used (in studies of a non-computational nature) in
cases of debated authorship, for example regarding some works
traditionally attributed to Dante Alighieri \citep{Hall:1989su,
Toynbee:1918ta}.

Given these premises, the goal of the present paper is to investigate
whether features based on syllabic quantity can be profitably employed
for computational AA in Latin prose texts. This seems reasonable also
for medieval Latin, since accents are heavily based on syllabic
quantity, as already explained. Features derived from syllabic
quantity are also content-agnostic (e.g., a sequence of syllables such
as ``long-short-long'' can stand for hundreds or thousands of
different $3$-syllable sequences), and thus they could be a valuable
tool for authorship analysis problems, since they avoid the risk of
unwanted influence from the domain.

% ----------------------------------------------------------------------

\subsection{Extracting syllabic quantity for Latin prose texts}
\label{sec:method_SQ_feats}

\noindent One important part of the computational system that needs to
be assembled in order to carry out our experiments on syllabic
quantity (from now on: SQ) is a module that, given a piece of Latin
written text, extracts SQ from it, in order to generate SQ-based
features that can be used for the classification task. Since
developing such a module would be a major endeavour, due to the
complexities of Latin prosody already mentioned in
Section~\ref{sec:method_Latin_prosody}, we decided to use an
off-the-shelf tool, chosen among those that are publicly available.

One such tool we considered is the one that resulted from the
\textit{Cursus in Clausula} project \citep[]{Spinazze:2014}: this tool
is a web application that not only extracts all the forms of
\textit{cursus} from an uploaded text, but it also allows performing
some statistical analyses on it. However, this tool analyses only the
final portions of periods and sentences (see the definitions of
\textit{clausula} and \textit{cursus} given in
Section~\ref{sec:method_Latin_prosody}). We believe instead that it
would be more profitable to extend our analysis to the entirety of the
document, and not only to the final portions of periods and sentences,
since this would highlight potential rhythmic preferences of an author
in creating the more general structure of the discourse; additionally,
we should remember that there are rhythmic rules also for the
beginning of the sentence \citep{Janson:1975po}.

In the end, we decided to implement our SQ extraction module by using
the Classical Language ToolKit library \citep{Johnson:2014lm}: among
many tools for the study of ancient languages, it offers specific ones
for the study of Latin prosody, in particular the two modules
\textsc{Macronizer} (which places a macron \={} over long vowels) and
\textsc{Scanner} (which produces a sequence of the traditional symbols
denoting the quantity of a syllable, i.e., short, long, and end of
sentence). The output of each functionality is shown in the respective
entry in Table~\ref{tab:cltk}.

\begin{table}[t]
  \centering
  \caption{Example results of the two modules from the CLTK library on
  a Latin prose text.}
  \label{tab:cltk}
  \begin{tabular}{|l|l|}
    \hline
    \textbf{Original text} & \begin{tabular}{l} Quo usque tandem abutere 
                               Catilina patientia nostra. \\ Quam diu etiam furor iste tuus nos 
                               eludet.
                             \end{tabular} 
    \\ \hline\hline
    \textbf{\textsc{Macronizer}} & \begin{tabular}{l} qu\={o} usque 
                                     tandem ab\={u}t\={e}re catil\={\i}na patientia nostra . \\ quam 
                                     di\={u} etiam furor iste tuus n\={o}s \={e}l\={u}det .\end{tabular} 
    \\ \hline
    \textbf{\textsc{Scanner}} & 
                                \begin{tabular}{l} $[-\cup-\cup--\cup\cup\cup-\cup\cup\cup-\cup\cup-X$, 
                                  \\ $-\cup\cup\cup-\cup\cup-\cup\cup----X]$
                                \end{tabular} \\ \hline
  \end{tabular}
\end{table}

% ----------------------------------------------------------------------

\section{Experimental setting}
\label{sec:exps}

\noindent As already stated, we focus specifically on the authorship
attribution (AA) problem.  Therefore, as mentioned in
Section~\ref{sec:intro}, given document $d$ we aim to assign it to
exactly one class among a set $\{A_{1}, ..., A_{m}\}$ of candidate
classes, i.e., possible authors. We evaluate the contribution of
SQ-derived features when added to other topic-agnostic features, using
different learning algorithms and different datasets; thus, the
quality of SQ-derived features is inferred from the difference between
the performance of a method without SQ-based features and the
performance of the same method while equipped with them.
 
In this section we outline the procedural details of our
experiments. In particular, we describe the datasets we have employed
(Section~\ref{sec:exps_datasets}), how we preprocess them
(Section~\ref{sec:preprocessing}), the topic-agnostic features we use
(Section~\ref{sec:exps_feats}), the different learning algorithms and
the evaluation protocol we use (Section~\ref{sec:exps_protocol}).

The programming language we employ in this project is Python; in
particular, we use several modules from the \texttt{scikit-learn}
\citep{Pedregosa:2011py} and \texttt{PyTorch} \citep{Paszke:2019th}
packages.

% ----------------------------------------------------------------------

\subsection{The datasets}
\label{sec:exps_datasets}

\noindent Currently, there is no Latin dataset that is considered a
standard in AA studies, and even the available ones are few.  In the
following sections we present the three Latin datasets we perform our
experiments on: one of them was assembled by us
(Section~\ref{sec:exps_datasets_LatinitasAntiqua}) while the other two
were originally presented in other AId works
(Section~\ref{sec:exps_datasets_KabalaCorpusA} and
Section~\ref{sec:exps_datasets_MedLatin}).

% ----------------------------------------------------------------------

\subsubsection{LatinitasAntiqua.}
\label{sec:exps_datasets_LatinitasAntiqua}

\noindent For the purpose of this work we have created an \textit{ad
hoc} Latin dataset to best suit our requirements.  For this we exploit
the \emph{Corpus Corporum} repository, developed by the University of
Zurich,\footnote{\url{http://www.mlat.uzh.ch/MLS/}} and in particular
its subsection called \textit{Latinitas Antiqua}, which contains
various Latin works from the Perseus Digital
library;\footnote{\url{http://www.perseus.tufts.edu/hopper/}} these
works are meticulously tagged in XML. From this section we further
select a group of texts that (a) are not poetry works, since our study
only deals with prose writings, and (b) are not theatrical pieces,
since these latter have a very peculiar format based on dialogue and
scenes.
% \footnote{It would certainly be interesting to verify the results of
% our SQ-based approach in an authorship analysis task addressed to
% theatrical works; we leave this for future investigation.}
The resulting dataset is presented in detail in
Table~\ref{tab:LatinitasAntiqua}. In total, it is composed of 90 prose
texts by 25 Latin authors, spanning the Classical, Imperial and Early
Medieval periods, and a variety of genres (mostly epistolary,
historical, and rhetoric).

By exploiting the XML tagging, from each text we delete foreign words
(e.g., Greek) and the direct quotations from other authors, in order
to retain only the ``pure'' production of the writer. We then remove
every remaining XML tag.

\begin{table}
  \centering
  \caption{The \textsf{LatinitasAntiqua} dataset. The table shows, for
  each author, the number of entire texts available (\#), as divided
  in the online repository, their titles, and the collective sum of
  words contained after the pre-processing of the data (see
  Section~\ref{sec:preprocessing}).}
  \label{tab:LatinitasAntiqua}
  % \fontsize{8}{9}\selectfont
  \resizebox{\textwidth}{!} {
  \begin{tabular}{|p{3.5cm}|r|m{8.0cm}|r|}
    \hline
    \multicolumn{1}{|c|}{\textbf{Author}} & 
                                            \multicolumn{1}{c|}{\textbf{\#}} & 
                                                                               \multicolumn{1}{c|}{\textbf{Titles}} & 
                                                                                                                      \multicolumn{1}{c|}{\textbf{Words}} 
    \\ 
    \hline
    Ammianus Marcellinus & 1 & \textit{Res gestae} & 121,751 \\ \hline
    Apuleius & 3 & \textit{Apologia, Florida, Metamorphoses} & 82,401 \\ \hline
    Augustinus Hipponensis & 1 & \textit{Epistularum Selectio} & 46,225 \\ \hline
    Aulus Gellius & 1 & \textit{Noctes Atticae} & 79,710 \\ \hline
    Beda & 1 & \textit{Historiam ecclesiasticam gentis Anglorum} & 59,238 \\ \hline
    Cicero & 29 & \textit{Academica, Brutus, Cato Maior: de Senectute, De Divinatione, De Fato, De Finibus, De Natura Deorum, De Officiis, De Optimo Genere Oratorum, De Oratore, De Republica, Epistolae ad Atticum, Epistolae ad Brutum, Epistolae ad Familiares, Epistolae ad Quintum, Laelius de Amicitia, Lucullus, Orationes (1, 2, 3, 4, 5, 6), Orator, Paradoxa stoicorum ad M. Brutum, Partitiones Oratoriae, Petitio consulatus, Topica, Tusculanae disputationes} & 1,078,559 \\ \hline
    Columella & 1 & \textit{De re rustica} & 78,782 \\ \hline
    Cornelius Celsus & 1 & \textit{De medicina} & 102,459 \\ \hline
    Cornelius Nepos & 1 & \textit{Vitae} & 28,377 \\ \hline
    Cornelius Tacitus & 5 & \textit{Annales, De Origine et Situ Germanorum Liber, De Vita Iulii Agricolae, Dialogus de Oratoribus, Historiae} & 161,937 \\ \hline
    Curtius Rufus & 1 & \textit{Historiae Alexandri Magni} & 74,260 \\ \hline Florus & 1 & \textit{Epitome Rerum Romanorum} & 26,237 \\ \hline Hieronymus Stridonensis & 1 & \textit{Epistulae Selectiones} & 43,692 \\ \hline Iulius Caesar & 2 & \textit{De Bello Civili, De Bello Gallico} & 83,628 \\ \hline Minucius Felix & 1 & \textit{Octavius} & 11,595 \\ \hline Plinius minor & 1 & \textit{Epistolae} & 64,881 \\ \hline Quintilianus & 1 & \textit{Institutio Oratoria} & 157,085 \\ \hline Sallustius & 2 & \textit{Bellum Iugurthinum, De Catilinae coniuratione} & 31,887 \\ \hline Seneca & 14 & \textit{Ad Lucilium Epistulae Morales, Apocolocyntosis, De Beneficiis, De Brevitate Vitae, De Clementia, De Consolatione ad Helvium, De Consolatione ad Marciam, De Consolatione ad Polybium, De Constantia, De Ira, De Otio, De Providentia, De Tranquillitate Animi, De Vita Beata} & 234,328 \\ \hline Seneca maior & 2 & \textit{Controversiae, Suasoriae} & 75,058 \\ \hline Servius & 3 & \textit{In Vergilii Aeneide comentarii, In Vergilii Eklogarum comentarii, In Vergilii Georgicis comentarii} & 290,921 \\ \hline Sidonius Apollinaris & 2 & \textit{Epistulae libri (1-7, 8-9)} & 45,583 \\ \hline Suetonius & 1 & \textit{De Vita Caesarum} & 69,764 \\ \hline Titus Livius & 13 & \textit{Ab Urbe Condita libri (1-2, 3-4, 5-7, 8-10, 21-22, 23-25, 26-27, 28-30, 31-34, 35-37, 38-39, 40-42, 43-45)} & 486,263 \\ \hline Vitruvius & 1 & \textit{De Architectura} & 57,221 \\ \hline
  \end{tabular} 
  }
\end{table}

% ----------------------------------------------------------------------

\subsubsection{KabalaCorpusA.}
\label{sec:exps_datasets_KabalaCorpusA}

\noindent This dataset was developed by \citet{Kabala:2020la}. In
particular, of the four datasets that he assembled, we exploit
\textsf{CorpusA}, the biggest one, which consists of $39$ texts by
$22$ authors from the 11th and 12th centuries. Long quotations and
passages of poetry have been already removed from the texts by the
author.

% ----------------------------------------------------------------------

\subsubsection{MedLatin.}
\label{sec:exps_datasets_MedLatin}

\noindent This dataset was developed by \citet{Corbara:2020tn}. The
authors originally divided it into two sub-datasets,
\textsf{MedLatinEpi} and \textsf{MedLatinLit}, both containing Latin
prose works mostly dating to the 13th and 14th centuries;
\textsf{MedLatinEpi} is composed of $294$ texts of epistolary genre,
while \textsf{MedLatinLit} is composed of $30$ texts of various
nature, especially comments on treatises and literary works. For this
project we combine the two sub-datasets together.\footnote{From
\textsf{MedLatinEpi} we exclude the texts from the collection of
Petrus de Boateriis, since the collection consists of a miscellanea of
different authors, often represented by just one epistle each; as
such, this collection is hardly useful for our goals.} We delete the
quotations from other authors and the parts in languages other than
Latin, both marked in the texts.

% ----------------------------------------------------------------------

\subsection{Preprocessing the data}
\label{sec:preprocessing}

\noindent We automatically pre-process all the documents in the three
datasets in order to clean them, as much as possible, from spurious
information and noise. In particular, we delete headings, editors'
notes, and other meta-information, if present. We delete symbols (such
as asterisks or parentheses) and Arabic numbers, since they are likely
bibliographical information inserted by the editor. We normalise
punctuation marks: we delete commas, and we replace all question
marks, exclamation marks, semicolons, colons and suspension points
with full periods. We do this because punctuation was absent or hardly
coherent in ancient manuscripts, hence the punctuation we see in
current editions follows modern habits, and is mostly due to the
editor, not to the author \citep[p.~57]{Tognetti:1982cr}. However, we
retain full periods in order to be able to divide the text into
sentences. We lowercase the text, and then we normalise it: we
exchange (i) all occurrences of character \textit{v} with character
\textit{u}, (ii) all occurrences of character \textit{j} with
character \textit{i}, and (iii) every stressed vowel with the
corresponding non-stressed vowel.\footnote{We do this in order to
standardise the different approaches that editors might follow. For
example, in medieval written Latin, instead of the two modern
graphemes ``u-U'' and ``v-V'', there was only one grapheme,
represented as a lowercase ``u'' and a capital ``V''; some
contemporary editors decide to follow this canon while others decide
to modernise the written text with the two separate graphemes.}

As a final step, we divide each text into sentences, where a sentence
is made of at least $5$ distinct words (we attach shorter sentences to
the next sentence in the sequence, or to the previous one in case the
sentence is the last one in the document). Each non-overlapping
sequence of 10 consecutive sentences is what we call a
\emph{fragment}. These fragments are the samples that we give as input
to our learning algorithms and classifiers; the final amount of
fragments for each of our three datasets is displayed in
Table~\ref{tab:datasets}.

\begin{table}[t]
  \centering
  \caption{Total of entire texts and resulting fragments in the
  datasets used in our experiments.}
  \label{tab:datasets}
  \begin{tabular}{l|r|r|r|}
    \cline{2-4}
    & \textsf{LatinitasAntiqua} & \textsf{KabalaCorpusA} & \textsf{MedLatin} \\ 
    \hline
    \multicolumn{1}{|l|}{\textbf{\# whole texts}} & 90 & 39 & 294 \\ 
    \hline
    \multicolumn{1}{|l|}{\textbf{\# fragments}} & 23,219 & 7,882 & 6,028 \\ 
    \hline
  \end{tabular}
\end{table}

% ----------------------------------------------------------------------

\subsection{Topic-agnostic features: base features and distorted
views}
\label{sec:exps_feats}

\noindent In surveying the results of the PAN 2011 shared task,
\citet{Argamon:2011ew} also describe the features that in 2011 were,
and have largely remained, standard for the representation of texts in
AId tasks. In the survey by \citet{Stamatatos:2009ye}, these features
are divided into five major types:
\begin{itemize}
\item \textbf{lexical}: features based on words and their patterns of
  occurrence in the text (e.g., word and sentence lengths, vocabulary
  richness, word $n$-grams, ...);
\item \textbf{character}: features based on characters and their
  patterns of occurrence in the text (e.g., character $n$-grams,
  compression measures, ...);
\item \textbf{syntactic}: features based on syntax (e.g.,
  part-of-speech tags, ...);
\item \textbf{semantic}: features based on semantics (e.g., synonyms,
  semantic dependencies, ...);
\item \textbf{application-specific}: features specifically engineered
  for the particular application under study (e.g., HTML tags, use of
  indentation, ...).
\end{itemize}

\noindent However, it has been frequently noted that certain AId
methods run a high risk of leveraging the domain (i.e., topic) the
text is about, rather than its style; in the terminology of
statistics, domain-dependent features here act as \textit{confounding
variables}. This means that, as pointed out for example in
\citep{Bischoff:2020mp, Halvani:2019sn}, if domain-dependent features
are used, an authorship classifier (even a seemingly good one) might
not really perform \textit{authorship} identification, as desired, but
might unintentionally perform \textit{topic} identification,
unwittingly leveraging not the stylistic peculiarities of an author
but the linguistic peculiarities typical of a certain topic.  Of
course, it is true that some authors confine their written production
to very restricted domains, but it would clearly be a poor decision to
classify a document as written by $A$ only because $A$ often or always
writes about the same topic the document is about. Word $n$-grams and
character $n$-grams may particularly suffer from this problem
\citep[]{Stamatatos:2009ye}; in fact, the good performance they
usually deliver in AId tasks may be due to the fact that the datasets
these tasks are tested on are often far from being topic-neutral.  It
would hence be good practice to avoid, as much as possible, using such
``topic markers'' when implementing authorship analysis algorithms.

With this goal in mind, various techniques can be employed. One
possibility consists of using only topic-agnostic features, such as
function words or syntactic features \citep{Jafariakinabad:2020sc,
Halvani2020:ee}.  A second possibility consists of actively
\textit{masking} topical content via a so-called ``text distortion''
approach \citep{Stamatatos:2018gm, Goot:2018bl}.  In order to obtain a
topic-agnostic representation of the text, we follow both routes
sketched above: we discuss them in Sections \ref{sec:BFs} and
\ref{sec:DVs}, respectively. We then assess the effect of SQ in AA
tasks by adding features derived from it to the topic-agnostic
representation of the text, using the difference in performance
between the two systems as a measure of said effect.

% ----------------------------------------------------------------------

\subsubsection{Base features.}
\label{sec:BFs}

\noindent We employ a set of features that are well-known and widely
used in the authorship analysis literature, and generally acknowledged
as being topic-independent. In this paper they will act as a common
base for each classifier, with other types of features being added to
them (according to a process based on the learning method used, see
Section~\ref{sec:exps_protocol_SVM} and
Section~\ref{sec:exps_protocol_NN}). We call this set
\textsc{BaseFeatures} (from now on: BFs); it is composed of the
following features:

\begin{itemize}
\item \textbf{function words}: the relative frequency of each function
  word. For a discussion about this type of features, see for example
  the study by \citet{Kestemont:2014uc}. We employ the following $80$
  Latin function words: \textit{a, ab, ac, ad, adhuc, ante, apud,
  atque, aut, autem, circa, contra, cum, de, dum, e, enim, ergo, et,
  etiam, ex, hec, iam, ibi, ideo, idest, igitur, in, inde, inter, ita,
  licet, nam, ne, nec, nisi, non, nunc, nunquam, ob, olim, per, post,
  postea, pro, propter, quando, quasi, que, quia, quidem, quomodo,
  quoniam, quoque, quot, satis, scilicet, sed, semper, seu, si, sic,
  sicut, sine, siue, statim, sub, super, supra, tam, tamen, tunc, ubi,
  uel, uelut, uero, uidelicet, unde, usque, ut}.
\item \textbf{word lengths}: the relative frequency of each possible
  word length, from a minimum of $1$ up to a maximum of $25$
  characters. These are standard features employed in statistical
  authorship analysis since Mendenhall's ``characteristic curves of
  composition'' \citep{Mendenhall:1887cc}.
\item \textbf{sentence lengths}: the relative frequency of each
  possible sentence length, from a minimum of $1$ up to a maximum of
  $100$ individual words. These features have been employed in
  statistical authorship analysis at least since the study of
  \citet{Yule:1939nt}.
\end{itemize}

\noindent
For each such feature type we compute a matrix $f\times t$, where $f$
is the number of fragments in the set and $t$ is the number of
features of the specific type, and we further scale each vector
to\textbf{} unit norm. Given the three resulting matrices, we
concatenate them in a single final matrix $f\times 205$, where $205$
is the total number of BFs.  We then do the same for the test set,
using the features extracted in the training phase. \silcorcomment{Non
capisco questa frase... Queste features sono 'fisse', cioè decise a
prescindere dal training set, non estratte da esso. Non sono come il
TfIdf, per esempio, per cui ovviamente contano solo i termini che si
trovano in training...} \fabsebcomment{Quello che intendevo è che $f$
è, immagino, il numero di frammenti nel \emph{training} set, quindi la
matrice che descrivi è quella dei documenti di training. Quindi, una
analoga matrice andrà fatta per i documenti di test. Sul discorso che
le features siano fisse o meno, be', immagino che le word lengths
siano da 1 a 26 perché 26 è la lunghezza della parola più lunga fra
quelle riscontrate nel training set (altrimenti non si capirebbe da
dove salta fuori questo 26; un numero magico?), quindi anche sul test
si usano word lengths da 1 a 26 (se, ad esempio, nel test set si
trovasse una parola lunga 27, questa non genererebbe una nuova
feature). Stessa cosa per le sentence lengths. O no?}

% ----------------------------------------------------------------------

\subsubsection{Distorted views.}
\label{sec:DVs}

\noindent We experiment with the four text ``distortion'' (i.e.,
masking) methods presented by \citet{Stamatatos:2018gm}, which aim to
preserve the document's stylistic characteristics while at the same
time hiding its topical content; each such method generates what
\citet{Stamatatos:2018gm} calls a \emph{distorted view} (from now on:
DV). Given a list $F$ of function words,\footnote{We employ the same
list of function words of Section~\ref{sec:BFs}.} the four DVs are:

\begin{itemize}
\item \textbf{Distorted View: Single Asterisk (DVSA)}: every word not
  included in $F$ is masked by replacing it with an asterisk ($*$).
\item \textbf{Distorted View: Multiple Asterisks (DVMA)}: every word
  not included in $F$ is masked by replacing each of its characters
  with an asterisk ($*$).
\item \textbf{Distorted View: Exterior Characters (DVEX)}: every word
  not included in $F$ is masked by replacing with an asterisk ($*$)
  each of its characters except the first and the last one. Note that
  one- and two-character words thus remain unaffected. As stated by
  \citet{Stamatatos:2018gm}, DVEX is based on many psychological
  studies underlining that ``exterior'' (i.e., first and last)
  characters are more important than ``interior'' characters for
  reading comprehension; thus, the positions of these characters
  appear to have a higher importance in how we process words (both in
  the reading activity and, one might imagine, in the writing
  activity). Additionally, the end of a word and the start of the
  following word might create sound effects that certain authors may
  want to avoid (e.g., using a word that begins with the same
  character as the end of the preceding word), or, conversely, to
  actively employ in their writing.
\item \textbf{Distorted View: Last 2 (DVL2)}: every word not included
  in $F$ is masked by replacing with an asterisk ($*$) each of its
  characters except the last two. Note that one- and two-character
  words thus remain unaffected.  Underlying DVL2 is the attempt to
  capture morpho-syntactic information (e.g., number, tense) that is
  often encoded in language via word suffixes.
\end{itemize}

\noindent The logic behind these masking methods is to remove any type
of topic-dependent information from the representation of the text,
while at the same time retaining topic-independent information. Some
of the information that is retained with these methods is independent
from word order, such as function words, word lengths, sentence
lengths, starting characters and ending characters of words, and their
frequencies; some of this information is already captured by the base
features of Section~\ref{sec:BFs}.  However, some of the information
that is retained is instead positional, i.e., dependent on word order;
examples are:
\begin{itemize}
\item for DVMA, DVEX, DVL2: the lengths of words that follow (or
  precede) specific function words, the lengths of words that are used
  as the first (or last) word of the sentence, the lengths of words
  that follow (or precede) short words (or long words), and their
  frequencies;
\item for DVEX: the frequencies with which a word begins (or ends)
  with certain characters, the frequencies with which a word that ends
  with a certain character is followed by a word that begins with
  another given character, etc.;
\item for DVL2: the frequencies with which a word ends with a given
  sequence of two characters, etc.

\end{itemize}
\noindent In other words,
% As noted in \citep{Stamatatos:2018gm}, DVMA allows taking into
% account the length of the single words, a kind of information that
% is lost in DVSA. However, opposed to the simple function words
% frequencies and word-lengths frequencies (which appear in the BFs
% set, see Section~\ref{sec:BFs}),
these DVs allow capturing the structure of the sentence, and this is
especially important when using learning mechanisms (such as the ones
of Section~\ref{sec:exps_protocol_NN}) to which the entire sequence is
given as input, and which are thus sensitive to positional
information.

% Note that, in the original DV methods, digits are replaced by one or
% more hashtags ($\#$); we skip this step, since we delete Arabic
% numbers in the pre-processing of the documents. Moreover, in the
% original study punctuation marks and capitalisation are preserved,
% unlike in our approach (for reasons already explained in
% Section~\ref{sec:preprocessing}).

By using the methods described in Section~\ref{sec:method_SQ_feats}
and in this section, we thus obtain five different ``encodings'' of
each document, i.e., the one representing SQ, and the four DVs
described by \citet{Stamatatos:2018gm}. From these five encodings we
can extract various kinds of features, an operation for which we also
need to take into account the specific learning algorithm we adopt;
this is unlike the set of BFs (see Section~\ref{sec:BFs}), since BFs
form a matrix which remains the same for both learning algorithms used
in this project. We discuss the feature extraction methods for SQ and
DVs in the sections specific to each learning method (Sections
\ref{sec:exps_protocol_SVM} and \ref{sec:exps_protocol_NN}). Note that
we also use the combination of the features extracted from all four
DVs; we call such a combination ALLDV.

% ----------------------------------------------------------------------

\subsection{Experimental protocol}
\label{sec:exps_protocol}

\noindent In this section we describe the protocol we adopt for the
experiments reported in this paper.

We assess the performance of the different classifiers on a given
dataset by randomly splitting the dataset into a training set
(containing 90\% of the data) and a test set (10\%). After performing
this split, we further remove from the training set 10\% of its data,
in order to use it as validation set.  This tri-partition of the
dataset into training set / validation set / test set is stratified,
meaning that the class distribution in the original dataset is
preserved in all three resulting subsets. For a given dataset, the
tri-partition is the same for all the systems being tested.

As the evaluation measure we use the well-known $F_{1}$ function,
i.e.,
\begin{equation}
  F_{1} = \frac{2\TP}{2\TP + 
  \FP + \FN}
\end{equation}
\noindent where $\TP$, $\FP$, and $\FN$ stand for the number of true
positives, false positives, and false negatives,
respectively. However, this is a measure for binary classification,
while our AA task is one of single-label multiclass
classification. Thus we compute \emph{macro-averaged $F_{1}$}
(hereafter: $F_{1}^{M}$) and \emph{micro-averaged $F_{1}$} (hereafter:
$F_{1}^{\mu}$), two evaluation measures for multiclass
classification. $F_{1}^{M}$ is computed by averaging the $F_{1}$
values obtained on each individual class (i.e., author), in a
one-vs-rest fashion (i.e., positive examples being those by the author
considered and negative examples being those by all other authors),
while $F_{1}^{\mu}$ is equal to the value of $F_{1}$ as obtained on
the aggregated contributions of all classes (i.e., where, say, FP is
the sum of the author-specific values of FP). Note that, in the case
of single-label multi-class classification, $F_{1}^{\mu}$ is
equivalent to the ``vanilla accuracy'' measure.

As anticipated, we aim to compute the difference in performance
between a method employing SQ-based features and the same method
without SQ-based features, using this difference as indicator of the
contribution of SQ to AA for Latin prose texts. To this aim, we also
compute the statistical significance of the aforementioned difference,
via McNemar's paired non-parametric statistical hypothesis test
\citep{McNemar:1947jh}. Since the test applies to binary results
(instead of categorical results), we convert the predictions of the
two methods of interest into binary values, where $1$ stands for a
correct prediction and $0$ stands for a wrong prediction. We take
$0.05$ as the confidence value for statistical significance.

In this project we experiment with two separate learning algorithms,
Support Vector Machines (SVMs) and ``deep'' Neural Networks (NNs); we
explain in detail the setup of both in the two sections that
follow. SVMs (Section~\ref{sec:exps_protocol_SVM}) are a standard
learning algorithm, widely used in text classification and, more
specifically, in AId (see also Section~\ref{sec:ml4aid}). As we will
show, in order to feed the five different encodings of the text (the
SQ-based encoding and the four DV-based encodings) to a SVM, we
extract character $n$-grams from the encoded texts. On the other hand,
our NN architecture (Section~\ref{sec:exps_protocol_NN}) does not
necessarily need this step, since it is possible to pass the encoded
texts to the neural network as they are.

% ----------------------------------------------------------------------

\subsubsection{Support vector machines.}
\label{sec:exps_protocol_SVM}

\noindent The SVM implementation we employ in this study is
\textsc{LinearSVC}, from the \texttt{scikit-learn}
package\footnote{\url{https://scikit-learn.org/}}
\citep{Pedregosa:2011py}. This implementation employs by default a
linear kernel and a one-vs-rest multi-class strategy, which has been
found by the developers similar in performance to the method by
\citet{Crammer:2001ti} (a standard algorithm for turning binary SVMs
into multiclass SVMs), but less demanding in terms of computational
cost.\footnote{See the documentation at
\url{https://scikit-learn.org/stable/modules/svm.html} for more
information.}

We experiment with various SVM-based classifiers, each characterized
by a specific feature set. Given the matrix of BFs (see
Section~\ref{sec:BFs}), which remains the same for all learning
methods, any additional feature set is simply concatenated to it, so
that the $f\times 205$ matrix of BFs becomes a matrix $f\times k$,
where ($k-205$) is the number of additional features, i.e., character
$n$-grams extracted from the various encodings of the text.

In particular, for the SQ encoding we use character $n$-grams (where a
``character'' ranges on the three SQ symbols ``$\cup$'', ``$-$'',
``$X$'') with all values of $n$ in the range $[\alpha, \beta]$. We set
$\alpha=3$ since many metric feet in Latin poetry are based on $3$
syllables, and we set $\beta=7$ because the most important cursus
rhythms are based on schemes between $5$ and $7$ syllables long (see
Section~\ref{sec:method_Latin_prosody}).  On the other hand, for each
of the DV encodings we follow \citet{Stamatatos:2018gm} and use
character $3$-grams that appear at least $5$ times in the training
set.

For all the features derived from the five encodings, we perform
feature weighting via TFIDF.\footnote{TFIDF is a method traditionally
used for feature weighting in information retrieval, i.e., for
attributing different degrees of importance to different features in
the same document vector \citep{Salton88}.
% It is based on two major assumptions: first, the more a term appears
% in a document, the more said term is relevant for the document (the
% \textit{term frequency} assumption); second, if a term is common in
% many documents in the collection, it is probably not particularly
% characterising for the single document (the \textit{inverse document
% frequency} assumption).
For the \texttt{scikit-learn} module we employ in order to compute
TFIDF, see
\url{https://scikit-learn.org/stable/modules/generated/sklearn.feature_extraction.text.TFIDFTransformer.html}.}
Since the SQ encoding gives rise to a large number of features (that
we will call \emph{SQ-grams}),\footnote{For instance, for the
\textsf{LatinitasAntiqua} dataset the SQ encoding gives rise to a
number of features one order of magnitude larger than either the DVSA
or the DVMA encodings.} we perform filter-style feature selection
(i.e., we retain the $p$ top-scoring features) using $\chi^{2}$
(probably the most frequently used feature selection function in
machine learning) as the feature scoring function.\footnote{We use the
$\chi^{2}$ implementation provided by \texttt{scikit-learn}, see
\url{https://scikit-learn.org/stable/modules/generated/sklearn.feature_selection.chi2.html}}
We do not perform feature selection on the BFs feature set and on the
$n$-grams extracted from the DVs encodings.
% via the \texttt{scikit-learn} module \texttt{SelectKBest}, which is
% based on the $\chi^{2}$ function.

We perform the optimisation of two parameters: the SVM parameter $C$,
which sets the trade-off between the training error and the margin,
and the feature selection factor $r$, which is the fraction of
SQ-grams that are retained as a result of the feature selection phase.
In particular, our approach is as follows:
\begin{enumerate}

\item We create a list of possible configurations for the classifier,
  where a configuration is made of a possible value for parameter $C$
  (we explore the range of values
  $[0.001, 0.01, 0.1, 1, 10, 100, 1000]$) and, if the method employs
  SQ-grams, a possible value for parameter $r$ (we explore the range
  of values $[0.1, 0.2,$ $0.3,$ $0.4,$ $0.5, 1.0]$). Thus, a possible
  configuration is $(C=10)$ if the method does not employ SQ-grams, or
  $(C=10,r=0.5$) if the method does.
\item For all configurations, we train a classifier with a certain
  configuration on the training set, and assess the performance of the
  classifier on the validation set.
\item Using the configuration that has scored the highest value of
  $F_{1}^{M}$ in validation, we train the final classifier from the
  union of training set and validation set.
\item We assess the final classifier on the test set, computing the
  values of $F_{1}^{M}$ and $F_{1}^{\mu}$.
\end{enumerate}

% ----------------------------------------------------------------------

\subsubsection{Neural networks.}
\label{sec:exps_protocol_NN}

\noindent The NN architecture for deep learning\footnote{For an
introduction to the terminology and theory of neural networks and deep
learning, see \citep{Goodfellow:2016ee}.} that we implement for this
work is based on the idea of having each of the $6$ feature types we
explore (1 based on the SQ-derived encoding, 4 based on the DV-derived
encodings, and 1 based on base features) work as a separate input
source for the network; this approach is inspired by the idea of the
multi-channel architecture developed by \citet{Ruder:2016jn}.
Specifically, the network consists of parallel input ``branches'',
each one processing a single feature type, with the outputs of the
parallel branches being combined in a single decision layer. Note that
the number of branches thus depends on the feature types one wants to
combine: in case of the matrix of BFs and a single encoding (for
example, only the SQ-based encoding), the network is made of only two
branches (one for BFs and one for the SQ-based encoding), and so
on. We implement the NN architecture with the \texttt{PyTorch} package
\citep{Paszke:2019th}.

The following setup for each input branch is the result of a series of
preliminary tests that we have run on the validation sets using
different combinations of layers. We have found that, for this task, a
simple sequence of Convolutional Neural Networks (CNNs) are more
effective than a combination of CNN and Long Short-Term Memory (LSTM)
layers, or a combination of CNN layers and an attention mechanism
\citep{Vaswani:2017tt}. This might be due to the limited size of the
datasets, which, when the number of parameters increases, does not
allow the network to properly generalise (we recall that NNs usually
require very large amounts of training data). Character-level CNNs
have been proven to work efficiently for text classification in
general \citep{Zhang:2015rc}, and on AA tasks in particular
\citep{Shrestha:2017cn}.
Each of the five encodings (from a document, i.e., a fragment) are
passed to the corresponding branch, and are independently padded so
that every string in the same input batch has the same length. A
branch is made of a total of 5 layers. Given $m$ classes (i.e.,
authors) and an input batch of $b$ fragments (we set the batch size to
$64$), the layers are as follows:
\begin{itemize}
\item \textbf{1 embedding layer.} It performs character embedding on
  the distorted text (the embedding size is set to $32$).
\item \textbf{2 CNN layers.} Each layer applies a convolutional
  operation with $128$ kernels of size $3$.  Afterward, a ReLU
  activation is applied, and shrinkage is performed on the output with
  a max-pooling operation. After the two layers, a dropout operation
  (with the drop probability set to $0.5$) is performed on the output.
\item \textbf{2 dense layers}. Each layer applies a ReLU operation
  over the input. Between the two layers, a dropout operation (with
  the drop probability set to $0.5$) is performed on the output. The
  final layer is made of $m$ neurons, i.e., the number of classes
  (i.e., authors).
\end{itemize}
\noindent A slightly different branch is devoted to the BFs, which are
passed as input in the form of a matrix (as already discussed in
Section~\ref{sec:BFs}). For this branch, the feature matrix is
processed by two dense layers with a ReLU activation, with a dropout
process in-between.

Hence, each branch returns a $b\times m$ matrix of probabilities,
where $b$ is the number of fragments in the input batch and $m$ is the
number of classes. These outputs are then stacked together, so as to
obtain a decision matrix of dimension $b\times m\times c$, where $c$
is the number of branches employed. An average-pooling operation is
applied on the dimension $c$, so that, for each class, the average
value of the decisions of the different branches is finally obtained,
yielding a $b\times m$ matrix. The final decisions, i.e., the author
classes predicted for the input fragments, are obtained via a final
dense layer which applies a softmax (for training) or argmax (for
testing) operation over the class probabilities.

\begin{figure}[t]
  \centering \includegraphics[width=\linewidth]{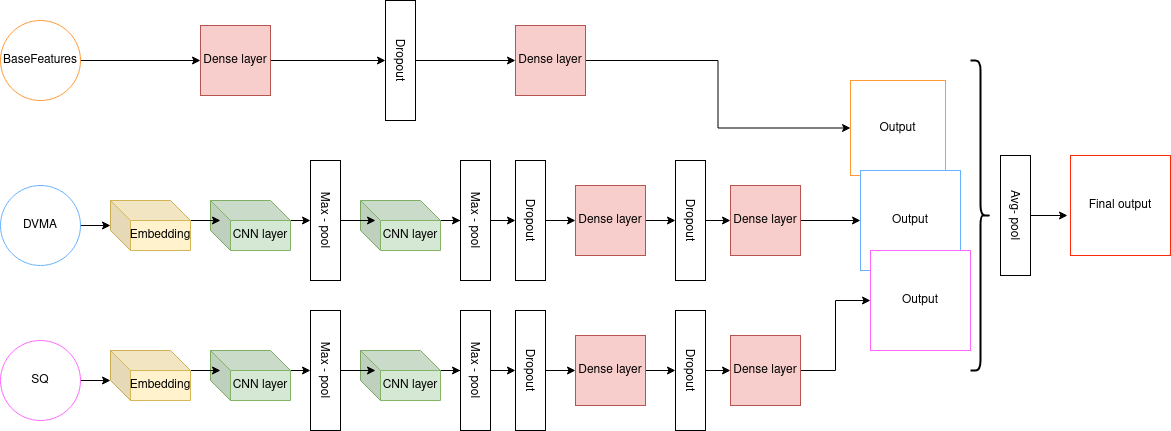}
  \caption{Scheme of the NN architecture developed; it is here
  instantiated with the BFs + DVMA + SQ setup.}
  \label{fig:NN}
\end{figure}

\citet{Menta:2021by} employ a similar approach in a SAV setting: in
their approach, two branches of different size process the information
coming from TFIDF-weighted (1) character $n$-grams and (2) punctuation
marks, after which the two outputs are combined into a single series
of layers that finally yields the classification result. Instead, in
our method each branch computes a separate classification decision,
and the decisions are then averaged. This is somewhat akin to
exploiting a committee (or ensemble) of classifiers, an approach which
has sometimes proven more effective than single classifiers
\citep{Adamovic:2019au, Potha:2019yc}. In the setting of AA,
\citet{Tearle:2008lk} have employed a similar committee of NNs, each
one with a different feature set; however, instead of hard-counting
the vote of each NN, as they do, we choose a ``softer'' approach and
average the probabilities computed by each branch, akin to what is
done by \citet{Muttenthaler:2019pa}. A scheme of our NN architecture
is displayed in Figure~\ref{fig:NN}.

The training of the NN is conducted with the traditional
backpropagation method; for this, we employ cross-entropy as the loss
function, and the Adam optimizer \citep{Kingma:2014ma}. In particular,
our approach is as follows:
\begin{enumerate}
\item We train the NN on the training set.
\item We assess the performance of the NN on the validation set,
  computing the $F_{1}^{M}$ value.
\item We repeat the previous steps for $100$ epochs without
  improvement before triggering an early stop (the maximum number of
  training epochs is $5000$, but, in the experiments we perform, early
  stopping always occurs after some hundreds epochs).
\item We train the model that has scored the highest $F_{1}^{M}$ in
  validation (before early stopping) with a final epoch on the
  validation set.
\item We assess the final result of the NN on the test set, computing
  the $F_{1}^{M}$ and $F_{1}^{\mu}$ values.
\end{enumerate}

\noindent A legitimate concern that may arise when comparing NN models
regards the possibility that the superiority of some methods (as
quantified via any evaluation measure) is due to the mere presence of
a higher number of parameters, and not to the supposedly more
informative content in the additional branches.  In order to check if
this is the case, we have developed a dummy channel named FAKE, acting
as control, which allows a fair comparison in terms of number of
parameters. This channel is different not in its architecture, but in
the masked text it processes: in fact, the text is masked with a
``fake'' version of the SQ encoding. Specifically, each word is
transformed as follows: given a number of syllables $s$ (which we
approximate to the number of characters divided by $3$), each syllable
is replaced by a random SQ symbol (``$\cup$'', ``$-$'', or
``$X$''). We collect all these transformations in a dictionary, so
that a certain word is always replaced by the same random sequence of
symbols, thus making the transformation uninformed but not equal to
random noise. The NN cannot learn anything useful from this kind of
information, hence the result obtained with this channel plays the
role of a baseline against which to compare the effect of the SQ
channel.

% ----------------------------------------------------------------------

\section{Results}
\label{sec:results}

\noindent As already mentioned, we perform two batches of experiments,
each with a different learning algorithm (SVMs and NNs). In each
batch, we compare various models with SQ-based features against the
same models without SQ-based features, in order to assess the
performance gain (if any) obtained by the addition of these
features. The results of the two batches of experiments are displayed
in Table~\ref{tab:results_svm_opt_second} (for the experiments based
on SVMs) and in Table~\ref{tab:results_nn} (for the experiments based
on NNs).

\begin{table}[t]
  \centering
  \caption{Results of the experiments with SVMs as the learning
  algorithm. Pairs of experiments, one without SQ-based features and
  the other with SQ-based features, are reported on two consecutive
  rows. For each experiment we report (i) the values $F_{1}^{M}$ and
  $F_{1}^{\mu}$, derived on a single train-validation-test split, (ii)
  the percentage improvement (indicated as $\Delta\%$) resulting from
  the addition of SQ-based features, and (iii) the results of the
  McNemar statistical significance test (\textbf{M}) against the
  baseline (\cmark : statistical significance confirmed; \xmark :
  statistical significance rejected). \textbf{Boldface} indicates the
  best result obtained for the given dataset and evaluation measure.}
  \label{tab:results_svm_opt_second}
  \resizebox{\columnwidth}{!}{%
  \begin{tabular}{l|rr|rr|c||rr|rr|c||rr|rr|c|}
    \cline{2-16}
    & \multicolumn{5}{c||}{\textsf{LatinitasAntiqua}} & \multicolumn{5}{c||}{\textsf{KabalaCorpusA}} & \multicolumn{5}{c|}{\textsf{MedLatin}} \\ \cline{2-16} 
    & \multicolumn{1}{c}{\textbf{$F_{1}^{M}$}} & \multicolumn{1}{c|}{\textbf{$\Delta\%$}} &\multicolumn{1}{c}{\textbf{$F_{1}^{\mu}$}} & \multicolumn{1}{c|}{\textbf{$\Delta\%$}} & \textbf{M} & \multicolumn{1}{c}{\textbf{$F_{1}^{M}$}} & \multicolumn{1}{c|}{\textbf{$\Delta\%$}} & \multicolumn{1}{c}{\textbf{$F_{1}^{\mu}$}} & \multicolumn{1}{c|}{\textbf{$\Delta\%$}} & \textbf{M} & \multicolumn{1}{c}{\textbf{$F_{1}^{M}$}} & \multicolumn{1}{c|}{\textbf{$\Delta\%$}} & \multicolumn{1}{c}{\textbf{$F_{1}^{\mu}$}} & \multicolumn{1}{c|}{\textbf{$\Delta\%$}} & \textbf{M} \\ \hline
    \multicolumn{1}{|l|}{BFs} & 0.626 &  & 0.713 &  &  & 0.616 &  & 0.654 &  &  & 0.691 &  & 0.791 &  &  \\
    \multicolumn{1}{|l|}{BFs + SQ} & 0.667 & +6.67 & 0.755 & +5.92 & \cmark & 0.645 & +4.69 & 0.672 & +2.71 & \xmark & 0.721 & +4.37 & 0.799 & +1.05 & \xmark \\ \hline
    \multicolumn{1}{|l|}{BFs + DVMA} & 0.608 &  & 0.725 &  &  & 0.644 &  & 0.689 &  &  & 0.767 &  & 0.829 &  &  \\
    \multicolumn{1}{|l|}{BFs + DVMA + SQ} & 0.722 & +18.78 & 0.795 & +9.68 & \cmark & 0.696 & +8.23 & 0.733 & +6.25 & \cmark & 0.735 & -4.26 & 0.818 & -1.40 & \xmark \\ \hline
    \multicolumn{1}{|l|}{BFs + DVSA} & 0.635 &  & 0.747 &  &  & 0.634 &  & 0.678 &  &  & 0.775 &  & 0.833 &  &  \\
    \multicolumn{1}{|l|}{BFs + DVSA + SQ} & 0.717 & +12.81 & 0.798 & +6.92 & \cmark & 0.687 & +8.43 & 0.728 & +7.29 & \cmark & 0.737 & -4.91 & 0.831 & -0.20 & \xmark \\ \hline
    \multicolumn{1}{|l|}{BFs + DVEX} & 0.829 &  & 0.880 &  &  & 0.816 &  & 0.831 &  &  & 0.844 &  & 0.902 &  &  \\
    \multicolumn{1}{|l|}{BFs + DVEX + SQ} & 0.844 & +1.89 & 0.887 & +0.83 & \xmark & 0.813 & -0.30 & 0.831 & +0.00 & \xmark & 0.857 & +1.53 & 0.910 & +0.92 & \xmark \\ \hline
    \multicolumn{1}{|l|}{BFs + DVL2} & 0.822 &  & 0.881 &  &  & 0.843 &  & 0.849 &  &  & 0.900 &  & 0.922 &  &  \\
    \multicolumn{1}{|l|}{BFs + DVL2 + SQ} & 0.855 & +3.92 & 0.896 & +1.71 & \cmark & 0.847 & +0.51 & 0.861 & +1.34 & \xmark & \textbf{0.927} & +3.00 & 0.935 & +1.44 & \xmark \\ \hline
    \multicolumn{1}{|l|}{BFs + ALLDV} & 0.871 &  & 0.918 &  &  & 0.856 &  & 0.863 &  &  & 0.845 &  & \textbf{0.944} &  &  \\
    \multicolumn{1}{|l|}{BFs + ALLDV + SQ} & \textbf{0.881} & +1.15 & \textbf{0.921} & +0.33 & \xmark & \textbf{0.883} & +3.08 & \textbf{0.882} & +2.20 & \cmark & 0.812 & -3.86 & 0.930 & -1.41 & \xmark \\ \hline
  \end{tabular}}
\end{table}

Regarding the experiments with SVMs, as can be seen in
Table~\ref{tab:results_svm_opt_second}, in most cases the accuracy of
the classifier is improved by the addition of the SQ-grams. This
effect is very substantial in \textsf{LatinitasAntiqua}, where the
presence of the SQ-grams always improves the accuracy, irrespectively
of classifier setup and evaluation measure, and in most cases does so
in a statistically significant sense. The improvement is smaller, and
often not statistically significant, in \textsf{KabalaCorpusA} and
\textsf{MedLatin}. We conjecture that this might be due to the fact
that rhythmic patterns are more discriminative when applied to authors
temporally distant from each other, since each of
\textsf{KabalaCorpusA} and \textsf{MedLatin} involve authors located
closer in time to each other. It is also worth noting that the
increased accuracy obtained through the use of SQ-grams tends to be
lower in methods employing the DVEX and DVL2 masking methods; this is
natural, since these methods reach very high $F_{1}$ values anyway,
which means that for them the margin of improvement is narrower.

\begin{table}[t]
  \centering
  \caption{Results of the experiments with NNs as the learning
  algorithm. Notational conventions are the same as for
  Table~\ref{tab:results_svm_opt_second}.}
  \label{tab:results_nn}
  \resizebox{\columnwidth}{!}{%
  \begin{tabular}{l|rr|rr|c||rr|rr|c||rr|rr|c|}
    \cline{2-16}
    & \multicolumn{5}{c||}{\textsf{LatinitasAntiqua}} & \multicolumn{5}{c||}{\textsf{KabalaCorpusA}} & \multicolumn{5}{c|}{\textsf{MedLatin}} \\ \cline{2-16} 
    & \multicolumn{1}{c}{\textbf{$F_{1}^{M}$}} & \multicolumn{1}{c|}{\textbf{$\Delta\%$}} &\multicolumn{1}{c}{\textbf{$F_{1}^{\mu}$}} & \multicolumn{1}{c|}{\textbf{$\Delta\%$}} & \textbf{M} & \multicolumn{1}{c}{\textbf{$F_{1}^{M}$}} & \multicolumn{1}{c|}{\textbf{$\Delta\%$}} & \multicolumn{1}{c}{\textbf{$F_{1}^{\mu}$}} & \multicolumn{1}{c|}{\textbf{$\Delta\%$}} & \textbf{M} & \multicolumn{1}{c}{\textbf{$F_{1}^{M}$}} & \multicolumn{1}{c|}{\textbf{$\Delta\%$}} & \multicolumn{1}{c}{\textbf{$F_{1}^{\mu}$}} & \multicolumn{1}{c|}{\textbf{$\Delta\%$}} & \textbf{M} \\ \hline
    \multicolumn{1}{|l|}{BFs} & 0.581 &  & 0.740 &  &  & 0.585 &  & 0.645 &  &  & 0.623 &  & 0.776 &  &  \\ 
    \multicolumn{1}{|l|}{BFs + SQ} & 0.585 & +0.63 & 0.740 & +0.00 & \xmark & 0.567 & -3.06 & 0.646 & +0.20 & \xmark  & 0.616 & -1.09 & 0.774 & -0.21 & \xmark \\ \hline
    % \multicolumn{1}{|l|}{BFs + FAKE} & 0.607 & +4.58 & 0.749 & +1.22
    % & \xmark & 0.527 & -9.87 & 0.630 & -2.36 & \xmark & 0.612 &
    % -1.77 & 0.766 & -1.28 & \xmark \\ \hline
    \multicolumn{1}{|l|}{BFs + DVMA} & 0.608 &  & 0.764 &  &  & 0.585 &  & 0.636 &  &  & 0.664 &  & 0.776 &  &  \\
    \multicolumn{1}{|l|}{BFs + DVMA + SQ} & 0.614 & +0.99 & 0.763 & -0.13 & \xmark & 0.554 & -5.30 & 0.636 & +0.00 & \xmark & 0.669 & +0.75 & 0.786 & +1.29 & \xmark \\ \hline
    \multicolumn{1}{|l|}{BFs + DVSA} & 0.592 &  & 0.758 &  &  & 0.555 &  & 0.625 &  &  & 0.674 &  & 0.788 &  &  \\
    \multicolumn{1}{|l|}{BFs + DVSA + SQ} & 0.647 & +9.29 & 0.772 & +1.85 & \xmark & 0.553 & -0.36 & 0.625 & +0.00 & \xmark & 0.652 & -3.26 & 0.769 & -2.41 & \xmark \\ \hline
    \multicolumn{1}{|l|}{BFs + DVEX} & 0.681 &  & 0.808 &  &  & 0.625 &  & 0.686 &  &  & 0.680 &  & 0.809 &  &  \\
    \multicolumn{1}{|l|}{BFs + DVEX + SQ} & 0.701 & +2.94 & 0.817 & +1.11 & \xmark & 0.615 & -1.60 & 0.681 & -0.73 & \xmark & 0.595 & -12.50 & 0.771 & -4.70 & \cmark \\ \hline
    \multicolumn{1}{|l|}{BFs + DVL2} & 0.671 &  & 0.800 &  &  & 0.646 &  & 0.697 &  &  & 0.710 &  & 0.818 &  &  \\
    \multicolumn{1}{|l|}{BFs + DVL2 + SQ} & 0.669 & -0.30 & 0.805 & +0.62 & \xmark & 0.605 & -6.35 & 0.687 & -1.43 & \xmark & 0.643 & -9.44 & 0.801 & -2.08 & \xmark \\ \hline
    \multicolumn{1}{|l|}{BFs + ALLDV} & \textbf{0.738} &  & 0.842 &  &  & 0.682 &  & 0.735 &  &  & 0.751 &  & 0.859 &  &  \\
    \multicolumn{1}{|l|}{BFs + ALLDV + SQ} & 0.732 & -0.81 & \textbf{0.844} & +0.24 & \xmark & \textbf{0.689} & +1.03 & \textbf{0.743} & +1.09 & \xmark & \textbf{0.753} & +0.27 & \textbf{0.866} & +0.81 & \xmark \\
    \multicolumn{1}{|l|}{BFs + ALLDV + FAKE} & 0.712 & -3.52 & 0.835 & -0.83 & \xmark & 0.671 & -1.61 & 0.731 & -0.54 & \xmark & 0.736 & -2.00 & 0.854 & -0.58 & \xmark \\ \hline
  \end{tabular}
  }
\end{table}

Regarding the batch of experiments based on NNs, whose results are
displayed in Table~\ref{tab:results_nn}, the effect of SQ-based
features on the computation is less clear, since the addition of these
features sometimes improves the accuracy and sometimes deteriorates
it, and since the difference in accuracy between a system equipped
with these features and the same system without them is almost never
(20 out of 21 times) statistically significant.

Note that we test the FAKE channel only in conjunction with all the DV
channels (i.e., in the ALLDV setup), since it is simply a control
experiment. In this experiment the FAKE channel performs worse than
the SQ channel, even generating a deterioration in performance with
respect to not including it. This clearly indicates that the effect of
the SQ channel on the network is not merely the result of a higher
number of parameters, but the result of some truly useful information.
% However, as already stated, the difference is almost never
% significant, thus such effect per se is a negligible one.

As in the SVM-based experiments, we witness some substantial
differences among datasets, with SQ-based features being more
beneficial on \textsf{LatinitasAntiqua} than on \textsf{KabalaCorpusA}
and \textsf{MedLatin}; this may be due to the fact that there are
substantially fewer training documents in \textsf{KabalaCorpusA} and
\textsf{MedLatin} than in \textsf{LatinitasAntiqua} (see
Table~\ref{tab:datasets}), which may mean that the former two datasets
do not have enough training information to exploit the power of
SQ-based features in a NN environment.

We observe that there may be not enough training data for the NN in a
more general sense. Note in fact that, for every combination of
feature sets, evaluation measure, and dataset, the SVM classifier
(Table~\ref{tab:results_svm_opt_second}) performs better (or much
better) than the NN classifier (Table~\ref{tab:results_nn}). This is
not surprising, since deep learning algorithms are notoriously
data-hungry.

In general, it is worth noting that,

\begin{itemize}

\item for 10 out of 12 combinations $\langle$learning algorithm,
  evaluation measure, dataset$\rangle$, the best-performing feature
  set involves SQ-based features;

\item for 7 combinations $\langle$learning algorithm, feature setup,
  dataset$\rangle$, SQ-based features bring about a statistically
  significant improvement in performance, while a statistically
  significant deterioration in performance due to the introduction of
  these features is observed only for 1 such combination.

\end{itemize}
\noindent Overall, this confirms that the idea of deriving rhythmic
features from syllabic quantity and to apply them to authorship
attribution for Latin prose text is a fruitful one.
% Moreover, although in the \textsf{LatinitasAntiqua} dataset the SQ
% channel almost always shows an improvement in the computation, in
% the other two datasets the reverse is true.  Nevertheless, it is
% worth to underline that the best-performing method for all the
% datasets is still always the BFs + ALLDV + SQ one (although in
% \textsf{LatinitasAntiqua} this is only true for the $F_{1}^{\mu}$
% metric).

% ----------------------------------------------------------------------

\section{Related work}
\label{sec:relatedwork}

\noindent In this section we provide an overview of past research
related to our project, starting from machine learning as applied to
AId tasks (Section~\ref{sec:ml4aid}), and moving on to AId tasks for
the Latin language (Section~\ref{sec:AId4latin}) and to AId works that
employ prosodic features (Section~\ref{sec:prosodic4AId}).

% ----------------------------------------------------------------------

\subsection{Machine learning for AId tasks}
\label{sec:ml4aid}

\noindent AId tasks are usually tackled by employing methods based on
machine learning (thereby viewing these tasks as instances of text
classification) or on distance metrics. The annual PAN shared tasks
(see e.g., \citep{Bevendorff:2020zw, Bevendorff:2021rt,
Kestemont:2019wo}) offer a very good overview of the most recent
trends in AId, often posing challenging problems in cross-domain
and/or open-set settings.

In particular, the baselines presented in the 2019 edition of PAN
\citep{Kestemont:2019wo} mirror the most frequently employed systems,
i.e., simple classifier-learning algorithms such as support vector
machines (SVMs) or logistic regression, distance functions based on
compression algorithms, and variations on the well-known Impostors
method \citep{Koppel:2014bq}. In particular, SVMs are a standard
learning algorithm for many text classification tasks, due to their
robustness to high dimensionalities and to their general
applicability. In various settings they are often shown to outperform
other learning algorithms such as decision trees and even neural
networks (NNs) \citep{Zheng:2006wf}.
% , even though the latter difference is reported as not statistically
% significant.

Despite the good results obtained in other natural language processing
tasks \citep{Young:2018tn}, for a long time NNs have rarely been
employed in AId tasks, arguably due to the huge quantity of training
data they usually require. Even though one of the first appearances of
NNs at PAN dates back to \citet{Bagnall:2015ro}, winner of the 2015
PAN edition \citep{Stamatatos:2015nv}, until recently it was generally
accepted that ``simple approaches based on character/word $n$-grams
and well-known classification algorithms are much more effective in
this task than more sophisticated methods based on deep learning''
\citep[p.9]{Kestemont:2018vo}. However, NN methods are nowadays
employed more and more frequently also in AId tasks
\citep{Bevendorff:2020zw, Bevendorff:2021rt}. In particular, NNs are
often employed to solve SAV tasks: it is a fairly common practice to
process both documents with the same NN, and compare the resulting
representations via a similarity measure (so that, if the final output
is lower than a certain threshold, the two documents are deemed
similar enough to be by the same author). This should force the NN to
find the best way to represent the documents, which in turn is a way
to maximise their stylistic differences when the documents are from
different authors (see for example the proposals by
\citet{Hosseinia:2018xs, Jasper:2018yh, Ding:2017lg}). A similar
notion is found in ``Siamese network'' architectures, where the two
documents are processed by parallel channels, and then presented to a
layer that performs a distance calculation (see for example the
proposals by
% \alexcomment{Qui forse potremmo aggiornare con altri articoli più
% recenti, come ``Siamese BERT for Authorship Verification'' di
% PAN2021}
\citet{Saedi:2021si, Araujo-Pino:2020se, Boenninghoff:2020pp}).
% \silcorcomment{Non riesco a trovare i proceedings di PAN2021, solo
% un extended abstract che spiega i task... Quindi forse non è il caso
% di citarne articoli?} \fabsebcomment{Chiedili a Potthast, o a Paolo
% Rosso, o a qualcuno di loro.}

% ----------------------------------------------------------------------

\subsection{AId for the Latin language}
\label{sec:AId4latin}

\noindent Beside applications in cybersecurity and forensics
\citep{Afroz:2014pp, Leonard:2017ci}, AId has also been used to help
philologists and literature scholars to untangle (or at least to
provide additional evidence for) some long-standing authorial
debates. Indeed, researchers have applied AId methods to historical
documents whose authorship has been lost, or hidden, during the
passing of centuries. In the present study we limit ourselves to the
application of these methods to the Latin language (both classical and
medieval), but many more cases have been studied in other languages,
starting with the seminal study of the Federalist Papers
\citep{Mosteller:1963fr}. However, ancient languages such as Latin
pose additional problems when compared to other (widely spoken) modern
languages, like the alterations due to the textual tradition and the
heavy limitations in data availability. Still, notable results have
been obtained in various studies, for example in the research on the
\textit{Corpus Caesarianum} \citep{Kestemont:2016ta} and on the
writing of Hildegard of Bingen \citep{Kestemont:2015ll}, in the
identification of Apuleius as the most probable author of a newly
found manuscript \citep{Stover:2016cl}, in the inquiry of
\citet{Tuccinardi:2017pp} regarding the authenticity of one of Pliny
the Younger’s letters, in the investigation of \citet{Kabala:2020la}
on the identities of the Monk of Lido and Gallus Anonymous, and in the
study regarding the so-called ``13th Epistle'' of Dante Alighieri
\citep{Corbara:2019cq, Corbara:2020tn}. Although most such researches
employ distance metrics or classical machine learning algorithms,
because of the already mentioned limited size of the available data,
the first experiments with neural approaches have also started to
appear, such as the CNN-based study of \citet{Vainio:2019gr}.

AId methodologies can be applied even to literary pieces whose
authorship is well-known and certain, in order to find possible
stylistic influences from other authors; for example, the goal of
\citet{Forstall:2011pu} is to verify a supposed influence by Catullus
on the poetry of Paul the Deacon.

% ----------------------------------------------------------------------

\subsection{Prosodic features in AId}
\label{sec:prosodic4AId}

\noindent In this landscape the idea of employing prosodic features is
not a new one. Of course, their most natural use is in studies focused
on poetry, such as in the study of \citet{Neidorf:2019rl} on the Old
English verse tradition, or in the already cited investigation by
\citet{Forstall:2011pu} on the supposed influence of Catullus on Paul
the Deacon's writings. Nevertheless, rhythmic or prosodic features
have also been employed in authorship analysis of prose text. However,
they usually consist of the study of word repetitions, like the
anaphora (the repetition of a word, or a sequence of words, from a
previous sentence at the beginning of a new sentence)
\citep{Lagutina:2021ml}, or they are based on mapping the texts into
the corresponding sequences of sounds before extracting the $n$-grams,
such as in the research by \citet{Forstall:2010fs}, where the authors
employ the CMU Pronouncing Dictionary for the
conversion.\footnote{Available at:
\url{http://www.speech.cs.cmu.edu/cgi-bin/cmudict}. As an example,
with this kind of conversion the word ``reason'' becomes ``R IY1 Z AH0
N''} Finally, syllables have been used as base units in other AId
works \citep{Sidorov:2018ci}, and more generally in other NLP tasks,
such as poem generation \citep{Zugarini:2019la}. These works, despite
being close in nature to the project we present here, explore a
linguistic dimension different from the one we aim to capture with the
study of syllabic quantity.

Some studies closer to ours employ the distribution of accent in order
to derive rhythmic features for AA in English. The work by
\citet{Dumalus:2011ta} is a pioneering one in this sense: using the
CMU Pronouncing Dictionary, they extract the pronunciation of each
word and transform it into a ``stress string'', where the symbols
$\{0,1,2\}$ represent the absence of stress, a primary stress, and a
secondary stress in the syllable, respectively.  \citet{Ivanov:2018ex}
improves on this work: since many English words are homographs (i.e.,
they have the same spelling but different pronunciation and meaning),
they select the correct pronunciation, and hence the correct stress
string, by studying the parts of speech of the words in the
text. Similarly, \citet{Plechavc:2021le} employs the frequencies of
``rhythmic types'' (where a rhythmic type is a bit string representing
the distribution of stressed and unstressed syllables in a line) as
features in tackling the attribution problem for \textit{Henry
VIII}. Accentuation, as explained in
Section~\ref{sec:method_Latin_prosody}, is related to the concept of
syllabic quantity (at least in the Latin language); however, to the
best of our knowledge, syllabic quantity has never been employed for
any AId tasks concerning prose texts.

% ----------------------------------------------------------------------

\section{Conclusion and future work}
\label{sec:conclusion}

\noindent In this project we exploit the notion of ``syllabic
quantity'' in order to derive features for the computational
authorship attribution of Latin prose texts; these features correspond
to sequences of syllables marked according to SQ, and are meant to
capture rhythmic aspects of textual discourse. In comparative
experiments over different datasets, different learners, and different
modes for representing topic-independent textual content, we show that
using SQ-derived information has a generally beneficial effect. In
particular, we experiment with three different Latin datasets and two
different learning algorithms, Support Vector Machines (SVMs) and
Neural Networks (NNs). SQ-derived information tends to have a more
beneficial effect when using SVMs than when using NNs, likely due to
the fact that NNs are notoriously demanding in terms of training data,
and that training data are not abundant in our datasets. For 5 out of
6 combinations of 2 evaluation measures $\times$ 3 datasets, the best
performance is obtained with a setup that involves SQ-derived
information. \silcorcomment{Forse è meglio ripetere quello che si dice
nei risultati (intendo, in quanto a formulazione)?}

% Especially in the SVM case, the increase in performance given by the
% SQ-based features is almost always statistically significant, while
% in the NN case their contribution is less clear, even though the
% best methods contain them in their feature space.

Future work along these lines might take at least three different
directions.
% First, we conjecture that masking the texts into the corresponding
% SQ might result into a fairly domain-free classification, due to the
% semantic agnosticism of a series of SQ. However, due to the time
% required to construct a Latin dataset labelled by topic or genre, we
% leave the proof for future work. \fabsebcomment{Non riesce a venirmi
% in mente a cosa potrbbe servire la classification by topic o by
% genre per i documenti antichi ...}
First, it would be interesting to see if the results we have obtained
on prose works are confirmed also on theatrical pieces, a type of text
that is not present in our datasets (and that we have deliberately
excluded from \textsf{LatinitasAntiqua} when creating it). In some
sense, this literary genre seems somehow intermediate between poetry
and prose, since it is rich in text of a declamatory nature; it is
thus conceivable that the analysis of SQ might be beneficial here too,
maybe even more than in the case of prose texts. Second, in this
project we focus on the AA task, but SQ-based features could easily be
employed in other authorship identification tasks, such as authorship
verification and same-authorship verification; it would be interesting
to check whether a beneficial effect of SQ-derived features is also
present in these other settings.  Third, and perhaps more important,
the fact that we have found SQ to have generally beneficial effects on
authorship attribution encourages to pursue the investigation of the
importance of rhythm on authorship identification tasks on languages
other than Latin, starting from ones linguistically close to Latin,
such as Italian or Spanish. In particular, it would be fascinating to
study whether modern-day prose writers unconsciously opt for specific
rhythmic patterns, to the point of being uniquely recognizable thanks
to them.
% As already hinted in Section~\ref{sec:method_Latin_prosody}, in
% modern languages the concept of syllabic quantity has almost been
% lost in favour of accentuation, and some works have tackled the task
% for English (see Section~\ref{sec:prosodic4AId}); however, there is
% still much room for research.

% ----------------------------------------------------------------------

\section*{Acknowledgments}
\label{sec:acks}

\noindent The first exploratory steps for this research have been
conducted during the preparation of the BSc thesis of Giulio
\citet{Canapa:2021ca}, co-supervised (aside from the 1st author and
3rd author) by Vittore Casarosa; we thank both Giulio and Vittore.  We
thank Mirko Tavoni for feedback on an earlier draft of this paper.

% -------------------------------------------------------------------

% \bibliographystyle{named} \bibliography{SyllabicQuantities,Silvia}

\begin{thebibliography}{}

\bibitem[\protect\citeauthoryear{Adamovic \bgroup \em et al.\egroup
  }{2019}]{Adamovic:2019au}
Sasa Adamovic, Vladislav Miskovic, Milan Milosavljevic, Marko Sarac, and Mladen
  Veinovic.
\newblock Automated language-independent authorship verification (for
  {Indo-European} languages).
\newblock {\em Journal of the Association for Information Science and
  Technology}, 70(8):858--871, 2019.

\bibitem[\protect\citeauthoryear{Afroz \bgroup \em et al.\egroup
  }{2014}]{Afroz:2014pp}
Sadia Afroz, Aylin~Caliskan Islam, Ariel Stolerman, Rachel Greenstadt, and
  Damon McCoy.
\newblock Doppelgänger finder: {Taking} stylometry to the underground.
\newblock In {\em Proceedings of the 35th IEEE Symposium on Security and
  Privacy (S\&P 2014)}, pages 212--226, Berkeley, US, 2014.

\bibitem[\protect\citeauthoryear{Allen \bgroup \em et al.\egroup
  }{1903}]{Ayer:2014lm}
Joseph~H. Allen, James~B. Greenough, George~L. Kittredge, and Albert~A. Howard,
  editors.
\newblock {\em {Allen} and {Greenough}’s new {Latin} grammar for schools and
  colleges}.
\newblock Ginn \& Company, Boston, US, 1903.

\bibitem[\protect\citeauthoryear{Araujo-Pino \bgroup \em et al.\egroup
  }{2020}]{Araujo-Pino:2020se}
Emir Araujo-Pino, Helena Gómez-Adorno, and Gibran~Fuentes Pineda.
\newblock Siamese network applied to authorship verification.
\newblock In {\em Working Notes of the 2020 Conference and Labs of the
  Evaluation Forum (CLEF 2020)}, Thessaloniki, GR, 2020.

\bibitem[\protect\citeauthoryear{Argamon and Juola}{2011}]{Argamon:2011ew}
Shlomo Argamon and Patrick Juola.
\newblock Overview of the international authorship identification competition
  at {PAN}-2011.
\newblock In {\em Notebook Papers of the CLEF 2011 Labs and Workshop},
  Amsterdam, {NL}, 2011.

\bibitem[\protect\citeauthoryear{Bagnall}{2015}]{Bagnall:2015ro}
Douglas Bagnall.
\newblock Author identification using multi-headed recurrent neural networks.
\newblock In {\em Working Notes of the 2015 Conference and Labs of the
  Evaluation Forum (CLEF 2015)}, Toulouse, FR, 2015.

\bibitem[\protect\citeauthoryear{Bevendorff \bgroup \em et al.\egroup
  }{2020}]{Bevendorff:2020zw}
Janek Bevendorff, Bilal Ghanem, Anastasia Giachanou, Mike Kestemont, Enrique
  Manjavacas, Ilia Markov, Maximilian Mayerl, Martin Potthast, Francisco
  M.~Rangel Pardo, Paolo Rosso, G{\"{u}}nther Specht, Efstathios Stamatatos,
  Benno Stein, Matti Wiegmann, and Eva Zangerle.
\newblock Overview of {PAN} 2020: {A}uthorship verification, celebrity
  profiling, profiling fake news spreaders on {T}witter, and style change
  detection.
\newblock In {\em Proceedings of the 2020 International Conference and Labs of
  the Evaluation Forum (CLEF 2020)}, pages 372--383, Thessaloniki, GR, 2020.

\bibitem[\protect\citeauthoryear{Bevendorff \bgroup \em et al.\egroup
  }{2021}]{Bevendorff:2021rt}
Janek Bevendorff, Berta Chulvi, Gretel Liz~De la~Pe{\~{n}}a~Sarrac{\'{e}}n,
  Mike Kestemont, Enrique Manjavacas, Ilia Markov, Maximilian Mayerl, Martin
  Potthast, Francisco Manuel~Rangel Pardo, Paolo Rosso, Efstathios Stamatatos,
  Benno Stein, Matti Wiegmann, Magdalena Wolska, and Eva Zangerle.
\newblock Overview of {PAN} 2021: {A}uthorship verification, profiling hate
  speech spreaders on {Twitter}, and style change detection (extended
  abstract).
\newblock In {\em Proceedings of the 43rd European Conference on Information
  Retrieval (ECIR 2021)}, pages 567--573, Lucca, IT, 2021.

\bibitem[\protect\citeauthoryear{Bischoff \bgroup \em et al.\egroup
  }{2020}]{Bischoff:2020mp}
Sebastian Bischoff, Niklas Deckers, Marcel Schliebs, Ben Thies, Matthias Hagen,
  Efstathios Stamatatos, Benno Stein, and Martin Potthast.
\newblock The importance of suppressing domain style in authorship analysis.
\newblock arXiv:2005.14714, 2020.

\bibitem[\protect\citeauthoryear{Boenninghoff \bgroup \em et al.\egroup
  }{2020}]{Boenninghoff:2020pp}
Benedikt Boenninghoff, Julian Rupp, Robert~M. Nickel, and Dorothea Kolossa.
\newblock Deep {Bayes} factor scoring for authorship verification.
\newblock In {\em Working Notes of the 2020 Conference and Labs of the
  Evaluation Forum (CLEF 2020)}, Thessaloniki, GR, 2020.

\bibitem[\protect\citeauthoryear{Canapa}{2021}]{Canapa:2021ca}
Giulio Canapa.
\newblock La quantità sillabica nella computational authorship attribution per
  testi latini.
\newblock BSc Thesis, University of Pisa, Pisa, IT, 2021.

\bibitem[\protect\citeauthoryear{Ceccarelli}{2018}]{Ceccarelli:2018so}
Lucio Ceccarelli.
\newblock {\em Prosodia e metrica latina classica, con cenni di metrica greca}.
\newblock Società Editrice Dante Alighieri, Roma, IT, 2018.

\bibitem[\protect\citeauthoryear{Corbara \bgroup \em et al.\egroup
  }{2019}]{Corbara:2019cq}
Silvia Corbara, Alejandro Moreo, Fabrizio Sebastiani, and Mirko Tavoni.
\newblock The {Epistle to Cangrande} through the lens of computational
  authorship verification.
\newblock In {\em Proceedings of the 1st International Workshop on Pattern
  Recognition for Cultural Heritage (PatReCH 2019)}, pages 148--158, Trento,
  IT, 2019.

\bibitem[\protect\citeauthoryear{Corbara \bgroup \em et al.\egroup
  }{2021}]{Corbara:2020tn}
Silvia Corbara, Alejandro Moreo, Fabrizio Sebastiani, and Mirko Tavoni.
\newblock {MedLatinEpi} and {MedLatinLit}: {Two} datasets for the computational
  authorship analysis of medieval {Latin} texts.
\newblock {\em ACM Journal of Computing and Cultural Heritage}, 2021.
\newblock Forthcoming.

\bibitem[\protect\citeauthoryear{Crammer and Singer}{2001}]{Crammer:2001ti}
Koby Crammer and Yoram Singer.
\newblock On the algorithmic implementation of multiclass kernel-based vector
  machines.
\newblock {\em Journal of Machine Learning Research}, 2(Dec):265--292, 2001.

\bibitem[\protect\citeauthoryear{Ding \bgroup \em et al.\egroup
  }{2017}]{Ding:2017lg}
Steven~H. Ding, Benjamin~C. Fung, Farkhund Iqbal, and William~K. Cheung.
\newblock Learning stylometric representations for authorship analysis.
\newblock {\em IEEE Transactions on Cybernetics}, 49(1):107--121, 2017.

\bibitem[\protect\citeauthoryear{Dumalus and Fernandez}{2011}]{Dumalus:2011ta}
Alvin~F. Dumalus and Proceso~L. Fernandez.
\newblock Authorship attribution using writer’s rhythm based on lexical
  stress.
\newblock In {\em Proceedings of the 11th Philippine Computing Science Congress
  (PCSC 2011)}, pages 82--88, Naga City, PH, 2011.

\bibitem[\protect\citeauthoryear{Forstall and Scheirer}{2010}]{Forstall:2010fs}
Chris Forstall and Walter Scheirer.
\newblock Features from frequency: {Authorship} and stylistic analysis using
  repetitive sound.
\newblock {\em Journal of the Chicago Colloquium on Digital Humanities and
  Computer Science}, 1(2), 01 2010.

\bibitem[\protect\citeauthoryear{Forstall \bgroup \em et al.\egroup
  }{2011}]{Forstall:2011pu}
Christopher~W. Forstall, Sarah~L. Jacobson, and Walter~J. Scheirer.
\newblock Evidence of intertextuality: {Investigating Paul the Deacon's
  Angustae Vitae}.
\newblock {\em Literary and Linguistic Computing}, 26(3):285--296, 2011.

\bibitem[\protect\citeauthoryear{Ginzburg}{1989}]{Ginzburg:1989ls}
Carlo Ginzburg.
\newblock {\em Clues: {Roots} of an {Evidential Paradigm}}, pages 96--125.
\newblock Johns Hopkins University Press, 1989.

\bibitem[\protect\citeauthoryear{Goodfellow \bgroup \em et al.\egroup
  }{2016}]{Goodfellow:2016ee}
Ian Goodfellow, Yoshua Bengio, and Aaron Courville.
\newblock {\em Deep learning}.
\newblock MIT press, Cambridge, US, 2016.

\bibitem[\protect\citeauthoryear{Hall and Sowell}{1989}]{Hall:1989su}
Ralph~G. Hall and Madison~U. Sowell.
\newblock `{Cursus}' in the {Can Grande} epistle: {A} forger shows his hand?
\newblock {\em Lectura Dantis}, (5):89--104, 1989.

\bibitem[\protect\citeauthoryear{Halvani \bgroup \em et al.\egroup
  }{2019}]{Halvani:2019sn}
Oren Halvani, Christian Winter, and Lukas Graner.
\newblock Assessing the applicability of authorship verification methods.
\newblock In {\em Proceedings of the 14th International Conference on
  Availability, Reliability and Security (ARES 2019)}, pages 1--10, Canterbury,
  UK, 2019.

\bibitem[\protect\citeauthoryear{Halvani \bgroup \em et al.\egroup
  }{2020}]{Halvani2020:ee}
Oren Halvani, Lukas Graner, and Roey Regev.
\newblock {TAVeer}: {An} interpretable topic-agnostic authorship verification
  method.
\newblock In {\em Proceedings of the 15th International Conference on
  Availability, Reliability and Security (ARES 2020)}, pages 1--10, Virtual
  Event, 2020.

\bibitem[\protect\citeauthoryear{Harrison}{2021}]{Harrison:web}
Rebecca Harrison.
\newblock Cogitatorium.
\newblock http://rharriso.sites.truman.edu/, 2021.
\newblock Accessed: 07-07-2021.

\bibitem[\protect\citeauthoryear{Hosseinia and
  Mukherjee}{2018}]{Hosseinia:2018xs}
Marjan Hosseinia and Arjun Mukherjee.
\newblock Experiments with {Neural Networks} for small and large scale
  authorship verification.
\newblock {\em arXiv preprint arXiv:1803.06456}, 2018.

\bibitem[\protect\citeauthoryear{Ivanov \bgroup \em et al.\egroup
  }{2018}]{Ivanov:2018ex}
Lubomir Ivanov, Amanda Aebig, and Stephen Meerman.
\newblock Lexical stress-based authorship attribution with accurate
  pronunciation patterns selection.
\newblock In {\em Proceedings of the 21st International Conference on Text,
  Speech, and Dialogue (TSD 2018)}, pages 67--75, Brno, CZ, 2018.

\bibitem[\protect\citeauthoryear{Jafariakinabad \bgroup \em et al.\egroup
  }{2020}]{Jafariakinabad:2020sc}
Fereshteh Jafariakinabad, Sansiri Tarnpradab, and Kien~A. Hua.
\newblock Syntactic neural model for authorship attribution.
\newblock In {\em Proceedings of the 33rd Conference of the Florida Artificial
  Intelligence Research Society (FLAIRS 2020)}, pages 234--239, Miami Beach,
  US, 2020.

\bibitem[\protect\citeauthoryear{Janson}{1975}]{Janson:1975po}
Tore Janson.
\newblock {\em Prose rhythm in medieval {Latin} from the 9th to the 13th
  century}.
\newblock Almqvist \& Wiksell International, Stockholm, SE, 1975.

\bibitem[\protect\citeauthoryear{Jasper \bgroup \em et al.\egroup
  }{2018}]{Jasper:2018yh}
Johannes Jasper, Philipp Berger, Patrick Hennig, and Christoph Meinel.
\newblock Authorship verification on short text samples using stylometric
  embeddings.
\newblock In {\em Proceedings of the 7th International Conference on Analysis
  of Images, Social Networks and Texts (AIST 2018)}, pages 64--75, Moscow,
  {RU}, 2018.

\bibitem[\protect\citeauthoryear{Johnson \bgroup \em et al.\egroup
  }{2021}]{Johnson:2014lm}
Kyle~P. Johnson, Patrick Burns, John Stewart, and Todd Cook.
\newblock {CLTK}: {The Classical Language Toolkit}, 2021.
\newblock https://github.com/cltk/cltk.

\bibitem[\protect\citeauthoryear{Juola}{2006}]{Juola:2006jn}
Patrick Juola.
\newblock Authorship attribution.
\newblock {\em Foundations and Trends in Information Retrieval},
  1(3):233–--334, 2006.

\bibitem[\protect\citeauthoryear{Kabala}{2020}]{Kabala:2020la}
Jakub Kabala.
\newblock Computational authorship attribution in medieval {Latin} corpora:
  {The} case of the {Monk of Lido} (ca. 1101--08) and {Gallus Anonymous} (ca.
  1113--17).
\newblock {\em Language Resources and Evaluation}, 54(1):25--56, 2020.

\bibitem[\protect\citeauthoryear{Keeline and Kirby}{2019}]{Keeline:2019au}
Tom Keeline and Tyler Kirby.
\newblock {`Auceps Syllabarum'}: {A} digital analysis of {Latin} prose rhythm.
\newblock {\em The Journal of Roman Studies}, 109:161--204, 2019.

\bibitem[\protect\citeauthoryear{Kestemont \bgroup \em et al.\egroup
  }{2015}]{Kestemont:2015ll}
Mike Kestemont, Sara Moens, and Jeroen Deploige.
\newblock Collaborative authorship in the twelfth century: {A} stylometric
  study of {Hildegard of Bingen and Guibert of Gembloux}.
\newblock {\em Digital Scholarship in the Humanities}, 30(2):199--224, 2015.

\bibitem[\protect\citeauthoryear{Kestemont \bgroup \em et al.\egroup
  }{2016}]{Kestemont:2016ta}
Mike Kestemont, Justin Stover, Moshe Koppel, Folgert Karsdorp, and Walter
  Daelemans.
\newblock Authenticating the writings of {Julius Caesar}.
\newblock {\em Expert Systems with Applications}, 63:86--96, 2016.

\bibitem[\protect\citeauthoryear{Kestemont \bgroup \em et al.\egroup
  }{2018}]{Kestemont:2018vo}
Mike Kestemont, Michael Tschuggnall, Efstathios Stamatatos, Walter Daelemans,
  Günther Specht, Benno Stein, and Martin Potthast.
\newblock Overview of the author identification task at {PAN-2018}:
  {Cross-domain} authorship attribution and style change detection.
\newblock In {\em Working Notes of the 2018 Conference and Labs of the
  Evaluation Forum (CLEF 2018)}, Avignon, FR, 2018.

\bibitem[\protect\citeauthoryear{Kestemont \bgroup \em et al.\egroup
  }{2019}]{Kestemont:2019wo}
Mike Kestemont, Efstathios Stamatatos, Enrique Manjavacas, Walter Daelemans,
  Martin Potthast, and Benno Stein.
\newblock Overview of the cross-domain authorship attribution task at {PAN}
  2019.
\newblock In {\em Working Notes of the 2019 Conference and Labs of the
  Evaluation Forum (CLEF 2019)}, Lugano, CH, 2019.

\bibitem[\protect\citeauthoryear{Kestemont}{2014}]{Kestemont:2014uc}
Mike Kestemont.
\newblock Function words in authorship attribution: {From} black magic to
  theory? \killpunct.
\newblock In {\em Proceedings of the 3rd Workshop on Computational Linguistics
  for Literature (CLFL 2014)}, pages 59--66, Göteborg, SE, 2014.

\bibitem[\protect\citeauthoryear{Kingma and Ba}{2015}]{Kingma:2014ma}
Diederik~P. Kingma and Jimmy Ba.
\newblock Adam: {A} method for stochastic optimization.
\newblock In {\em Proceedings of the 3rd International Conference on Learning
  Representations (ICLR 2015)}, San Diego, US, 2015.

\bibitem[\protect\citeauthoryear{Koppel and Winter}{2014}]{Koppel:2014bq}
Moshe Koppel and Yaron Winter.
\newblock Determining if two documents are written by the same author.
\newblock {\em Journal of the Association for Information Science and
  Technology}, 65(1):178--187, 2014.

\bibitem[\protect\citeauthoryear{Koppel \bgroup \em et al.\egroup
  }{2009}]{Koppel:2009ix}
Moshe Koppel, Jonathan Schler, and Shlomo Argamon.
\newblock Computational methods in authorship attribution.
\newblock {\em Journal of the American Society for Information Science and
  Technology}, 60(1):9--26, 2009.

\bibitem[\protect\citeauthoryear{Lagutina \bgroup \em et al.\egroup
  }{2021}]{Lagutina:2021ml}
Ksenia Lagutina, Nadezhda Lagutina, Elena Boychuk, Vladislav Larionov, and Ilya
  Paramonov.
\newblock Authorship verification of literary texts with rhythm features.
\newblock In {\em Proceedings of the 28th Conference of the Finnish-Russian
  University Cooperation in Telecommunications (FRUCT 2021)}, pages 240--251,
  Oulu, FI, 2021.

\bibitem[\protect\citeauthoryear{Leonard \bgroup \em et al.\egroup
  }{2017}]{Leonard:2017ci}
Robert~A. Leonard, Juliane~E. Ford, and Tanya~K. Christensen.
\newblock Forensic linguistics: {Applying} the science of linguistics to issues
  of the law.
\newblock {\em Hofstra Law Review}, 45(3):11, 2017.

\bibitem[\protect\citeauthoryear{McNemar}{1947}]{McNemar:1947jh}
Quinn McNemar.
\newblock Note on the sampling error of the difference between correlated
  proportions or percentages.
\newblock {\em Psychometrika}, 12(2):153--157, 1947.

\bibitem[\protect\citeauthoryear{Mendenhall}{1887}]{Mendenhall:1887cc}
Thomas~C. Mendenhall.
\newblock The characteristic curves of composition.
\newblock {\em Science}, 9(214):237--249, 1887.

\bibitem[\protect\citeauthoryear{Menta and Garcia-Serrano}{2021}]{Menta:2021by}
Antonio Menta and Ana Garcia-Serrano.
\newblock Authorship verification with neural networks via stylometric feature
  concatenation.
\newblock In {\em Working Notes of the Conference and Labs of the Evaluation
  Forum (CLEF 2021)}, Bucharest, RO, 2021.

\bibitem[\protect\citeauthoryear{Mosteller and
  Wallace}{1963}]{Mosteller:1963fr}
Frederick Mosteller and David~L Wallace.
\newblock Inference in an authorship problem: {A} comparative study of
  discrimination methods applied to the authorship of the disputed {Federalist
  Papers}.
\newblock {\em Journal of the American Statistical Association},
  58(302):275--309, 1963.

\bibitem[\protect\citeauthoryear{Muttenthaler \bgroup \em et al.\egroup
  }{2019}]{Muttenthaler:2019pa}
Lukas Muttenthaler, Gordon Lucas, and Janek Amann.
\newblock Authorship attribution in fan-fictional texts given variable length
  character and word n-grams.
\newblock In {\em Working Notes of the 2019 Conference and Labs of the
  Evaluation Forum (CLEF 2019)}, Lugano, CH, 2019.

\bibitem[\protect\citeauthoryear{Neidorf \bgroup \em et al.\egroup
  }{2019}]{Neidorf:2019rl}
Leonard Neidorf, Madison~S. Krieger, Michelle Yakubek, Pramit Chaudhuri, and
  Joseph~P. Dexter.
\newblock Large-scale quantitative profiling of the {Old English} verse
  tradition.
\newblock {\em Nature Human Behaviour}, 3(6):560--567, 2019.

\bibitem[\protect\citeauthoryear{Oberhelman and Hall}{1984}]{Oberhelman:1984nw}
Steven~M. Oberhelman and Ralph~G. Hall.
\newblock A new statistical analysis of accentual prose rhythms in imperial
  {Latin} authors.
\newblock {\em Classical Philology}, 79(2):114--130, 1984.

\bibitem[\protect\citeauthoryear{Paszke \bgroup \em et al.\egroup
  }{2019}]{Paszke:2019th}
Adam Paszke, Sam Gross, Francisco Massa, Adam Lerer, James Bradbury, Gregory
  Chanan, Trevor Killeen, Zeming Lin, Natalia Gimelshein, Luca Antiga, Alban
  Desmaison, Andreas Kopf, Edward Yang, Zachary DeVito, Martin Raison, Alykhan
  Tejani, Sasank Chilamkurthy, Benoit Steiner, Lu~Fang, Junjie Bai, and Soumith
  Chintala.
\newblock Pytorch: {An} imperative style, high-performance deep learning
  library.
\newblock In {\em Proceedings of the 33rd Conference on Neural Information
  Processing Systems (NeurIPS 2019)}, pages 8024--8035, Vancouver, {CA}, 2019.

\bibitem[\protect\citeauthoryear{Pedregosa \bgroup \em et al.\egroup
  }{2011}]{Pedregosa:2011py}
Fabian Pedregosa, Gaël Varoquaux, Alexandre Gramfort, Vincent Michel, Bertrand
  Thirion, Olivier Grisel, Mathieu Blondel, Peter Prettenhofer, Ron Weiss,
  Vincent Dubourg, Jake Vanderplas, Alexandre Passos, David Cournapeau,
  Matthieu Brucher, Matthieu Perrot, and Édouard Duchesnay.
\newblock Scikit-learn: {Machine} learning in {P}ython.
\newblock {\em Journal of Machine Learning Research}, 12:2825--2830, 2011.

\bibitem[\protect\citeauthoryear{Plechá{\v{c}}}{2021}]{Plechavc:2021le}
Petr Plechá{\v{c}}.
\newblock Relative contributions of {Shakespeare} and {Fletcher} in {Henry
  VIII}: {An} analysis based on most frequent words and most frequent rhythmic
  patterns.
\newblock {\em Digital Scholarship in the Humanities}, 36(2):430--438, 2021.

\bibitem[\protect\citeauthoryear{Potha and Stamatatos}{2019}]{Potha:2019yc}
Nektaria Potha and Efstathios Stamatatos.
\newblock Dynamic ensemble selection for author verification.
\newblock In {\em Proceedings of the 41st European Conference on Information
  Retrieval (ECIR 2019)}, pages 102--115, Köln, DE, 2019.

\bibitem[\protect\citeauthoryear{Ruder \bgroup \em et al.\egroup
  }{2016}]{Ruder:2016jn}
Sebastian Ruder, Parsa Ghaffari, and John~G. Breslin.
\newblock Character-level and multi-channel convolutional neural networks for
  large-scale authorship attribution.
\newblock arXiv:1609.06686, 2016.

\bibitem[\protect\citeauthoryear{Saedi and Dras}{2021}]{Saedi:2021si}
Chakaveh Saedi and Mark Dras.
\newblock Siamese networks for large-scale author identification.
\newblock {\em Computer Speech \& Language}, 70:101241, 2021.

\bibitem[\protect\citeauthoryear{Salton and Buckley}{1988}]{Salton88}
Gerard Salton and Christopher Buckley.
\newblock Term-weighting approaches in automatic text retrieval.
\newblock {\em Information Processing and Management}, 24(5):513--523, 1988.

\bibitem[\protect\citeauthoryear{Shrestha \bgroup \em et al.\egroup
  }{2017}]{Shrestha:2017cn}
Prasha Shrestha, Sebastian Sierra, Fabio~A. González, Manuel Montes, Paolo
  Rosso, and Thamar Solorio.
\newblock Convolutional neural networks for authorship attribution of short
  texts.
\newblock In {\em Proceedings of the 15th Conference of the European Chapter of
  the Association for Computational Linguistics (EACL 2017)}, pages 669--674,
  Valencia, {ES}, 2017.

\bibitem[\protect\citeauthoryear{Sidorov}{2018}]{Sidorov:2018ci}
Grigori~O. Sidorov.
\newblock Automatic authorship attribution using syllables as classification
  features.
\newblock {\em Rhema}, (1):62--81, 2018.

\bibitem[\protect\citeauthoryear{Spinazzè}{2014}]{Spinazze:2014}
Linda Spinazzè.
\newblock {`Cursus in Clausula'}: {A}n online analysis tool of {Latin} prose.
\newblock In {\em Proceedings of the 3rd AIUCD Annual Conference on Humanities
  and Their Methods in the Digital Ecosystem}, pages 1--6, Bologna, IT, 2014.

\bibitem[\protect\citeauthoryear{Stamatatos \bgroup \em et al.\egroup
  }{2015}]{Stamatatos:2015nv}
Efstathios Stamatatos, Walter Daelemans, Ben Verhoeven, Patrick Juola, Aurelio
  López-López, Martin Potthast, and Benno Stein.
\newblock Overview of the author identification task at {PAN 2015}.
\newblock In {\em Working Notes of the 2015 Conference and Labs of the
  Evaluation Forum (CLEF 2015)}, Toulouse, {FR}, 2015.

\bibitem[\protect\citeauthoryear{Stamatatos}{2009}]{Stamatatos:2009ye}
Efstathios Stamatatos.
\newblock A survey of modern authorship attribution methods.
\newblock {\em Journal of the American Society for information Science and
  Technology}, 60(3):538--556, 2009.

\bibitem[\protect\citeauthoryear{Stamatatos}{2016}]{Stamatatos:2016ij}
Efstathios Stamatatos.
\newblock Authorship verification: {A} review of recent advances.
\newblock {\em Research in Computing Science}, 123:9--25, 2016.

\bibitem[\protect\citeauthoryear{Stamatatos}{2018}]{Stamatatos:2018gm}
Efstathios Stamatatos.
\newblock Masking topic-related information to enhance authorship attribution.
\newblock {\em Journal of the Association for Information Science and
  Technology}, 69(3):461--473, 2018.

\bibitem[\protect\citeauthoryear{Stover \bgroup \em et al.\egroup
  }{2016}]{Stover:2016cl}
Justin~Anthony Stover, Yaron Winter, Moshe Koppel, and Mike Kestemont.
\newblock Computational authorship verification method attributes a new work to
  a major 2nd century {African} author.
\newblock {\em Journal of the Association for Information Science and
  Technology}, 67(1):239--242, 2016.

\bibitem[\protect\citeauthoryear{Sturtevant}{1922}]{Sturtevant:1922ll}
Edgar~Howard Sturtevant.
\newblock Syllabification and syllabic quantity in {Greek} and {Latin}.
\newblock {\em Transactions and Proceedings of the American Philological
  Association}, 53:35--51, 1922.

\bibitem[\protect\citeauthoryear{Tearle \bgroup \em et al.\egroup
  }{2008}]{Tearle:2008lk}
Matt Tearle, Kye Taylor, and Howard Demuth.
\newblock An algorithm for automated authorship attribution using neural
  networks.
\newblock {\em Literary and Linguistic Computing}, 23(4):425--442, 2008.

\bibitem[\protect\citeauthoryear{Tognetti}{1982}]{Tognetti:1982cr}
Giampaolo Tognetti.
\newblock {\em Criteri per la trascrizione di testi medievali latini e
  italiani}, volume~51 of {\em Quaderni della Rassegna degli Archivi di Stato}.
\newblock Ministero per i beni culturali e ambientali, Roma, IT, 1982.

\bibitem[\protect\citeauthoryear{Toynbee}{1918}]{Toynbee:1918ta}
Paget Toynbee.
\newblock {Dante} and the {Cursus}: {A} new argument in favour of the
  authenticity of the '{Quaestio de Aqua et Terra}'.
\newblock {\em The Modern Language Review}, 13(4):420--430, 1918.

\bibitem[\protect\citeauthoryear{Tuccinardi}{2017}]{Tuccinardi:2017pp}
Enrico Tuccinardi.
\newblock An application of a profile-based method for authorship verification:
  {Investigating} the authenticity of {Pliny the Younger}'s letter to {Trajan}
  concerning the {Christians}.
\newblock {\em Digital Scholarship in the Humanities}, 32(2):435--447, 2017.

\bibitem[\protect\citeauthoryear{Vainio \bgroup \em et al.\egroup
  }{2019}]{Vainio:2019gr}
Raija Vainio, Reima Välimäki, Anni Hella, Marjo Kaartinen, Teemu Immonen,
  Aleksi Vesanto, and Filip Ginter.
\newblock Reconsidering authorship in the {Ciceronian} corpus through
  computational authorship attribution.
\newblock {\em Ciceroniana On Line}, 3(1):15--48, 2019.

\bibitem[\protect\citeauthoryear{van~der Goot \bgroup \em et al.\egroup
  }{2018}]{Goot:2018bl}
Rob van~der Goot, Nikola Ljube{\v{s}}ić, Ian Matroos, Malvina Nissim, and
  Barbara Plank.
\newblock Bleaching text: {Abstract} features for cross-lingual gender
  prediction.
\newblock In {\em Proceedings of the 56th Annual Meeting of the Association for
  Computational Linguistics (ACL 2018), Volume 2: Short Papers}, pages
  383--389, Melbourne, AU, 2018.

\bibitem[\protect\citeauthoryear{Vaswani \bgroup \em et al.\egroup
  }{2017}]{Vaswani:2017tt}
Ashish Vaswani, Noam Shazeer, Niki Parmar, Jakob Uszkoreit, Llion Jones,
  Aidan~N. Gomez, Lukasz Kaiser, and Illia Polosukhin.
\newblock Attention is all you need.
\newblock In {\em Proceedings of the 31st Conference on Neural Information
  Processing Systems (NIPS 2017)}, pages 5998--–6008, Long Beach, US, 2017.

\bibitem[\protect\citeauthoryear{Young \bgroup \em et al.\egroup
  }{2018}]{Young:2018tn}
Tom Young, Devamanyu Hazarika, Soujanya Poria, and Erik Cambria.
\newblock Recent trends in deep-learning-based natural language processing.
\newblock {\em IEEE Computational Intelligence}, 13(3):55--75, 2018.

\bibitem[\protect\citeauthoryear{Yule}{1939}]{Yule:1939nt}
G.~Udny Yule.
\newblock On sentence-length as a statistical characteristic of style in prose:
  {With} application to two cases of disputed authorship.
\newblock {\em Biometrika}, 30(3/4):363--390, 1939.

\bibitem[\protect\citeauthoryear{Zhang \bgroup \em et al.\egroup
  }{2015}]{Zhang:2015rc}
Xiang Zhang, Junbo Zhao, and Yann LeCun.
\newblock Character-level convolutional networks for text classification.
\newblock In {\em Proceedings of the 29th Annual Conference on Neural
  Information Processing Systems (NIPS 2015)}, pages 649–--657, Montreal,
  {CA}, 2015.

\bibitem[\protect\citeauthoryear{Zheng \bgroup \em et al.\egroup
  }{2006}]{Zheng:2006wf}
Rong Zheng, Jiexun Li, Hsinchun Chen, and Zan Huang.
\newblock A framework for authorship identification of online messages:
  {Writing}-style features and classification techniques.
\newblock {\em Journal of the American Society for Information Science and
  Technologies}, 57(3):378--393, 2006.

\bibitem[\protect\citeauthoryear{Zugarini \bgroup \em et al.\egroup
  }{2019}]{Zugarini:2019la}
Andrea Zugarini, Stefano Melacci, and Marco Maggini.
\newblock Neural poetry: {Learning} to generate poems using syllables.
\newblock In {\em Proceedings of the 28th International Conference on
  Artificial Neural Networks (ICANN 2019)}, pages 313--325, Munich, DE, 2019.

\end{thebibliography}

% -------------------------------------------------------------------

\end{document}